\documentclass[10pt,twocolumn,letterpaper]{article}

\usepackage{cvpr}
\usepackage{times}
\usepackage{epsfig}
\usepackage{graphicx}
\usepackage{amsmath}
\usepackage{amssymb}
\usepackage{bm}
\usepackage{makecell}
\usepackage{setspace}
\usepackage{array}
\newcolumntype{P}[1]{>{\centering\arraybackslash}p{#1}}

\makeatletter
\renewcommand{\paragraph}{
  \@startsection{paragraph}{4}
                {\z@}{1ex \@plus 0ex \@minus .0ex}{-1em}
                {\normalfont\normalsize\bfseries}
}
\makeatother

\newcommand{\bhline}[1]{\noalign{\hrule height #1}}

\usepackage[pagebackref=true,breaklinks=true,letterpaper=true,colorlinks,bookmarks=false]{hyperref}

\cvprfinalcopy

\begin{document}

\title{Generative Adversarial Image Synthesis with Decision Tree Latent Controller}

\author{
  Takuhiro Kaneko \hspace{7mm}
  Kaoru Hiramatsu \hspace{7mm}
  Kunio Kashino\\
  NTT Communication Science Laboratories, NTT Corporation\\
  {\tt\small \{kaneko.takuhiro, hiramatsu.kaoru, kashino.kunio\}@lab.ntt.co.jp}
}

\maketitle

\begin{abstract}
  This paper proposes the decision tree latent controller generative adversarial network (DTLC-GAN), an extension of a GAN that can learn hierarchically interpretable representations without relying on detailed supervision. To impose a hierarchical inclusion structure on latent variables, we incorporate a new architecture called the DTLC into the generator input. The DTLC has a multiple-layer tree structure in which the ON or OFF of the child node codes is controlled by the parent node codes. By using this architecture hierarchically, we can obtain the latent space in which the lower layer codes are selectively used depending on the higher layer ones. To make the latent codes capture salient semantic features of images in a hierarchically disentangled manner in the DTLC, we also propose a hierarchical conditional mutual information regularization and optimize it with a newly defined curriculum learning method that we propose as well.  This makes it possible to discover hierarchically interpretable representations in a layer-by-layer manner on the basis of information gain by only using a single DTLC-GAN model. We evaluated the DTLC-GAN on various datasets, i.e., MNIST, CIFAR-10, Tiny ImageNet, 3D Faces, and CelebA, and confirmed that the DTLC-GAN can learn hierarchically interpretable representations with either unsupervised or weakly supervised settings. Furthermore, we applied the DTLC-GAN to image-retrieval tasks and showed its effectiveness in representation learning.
\end{abstract}

\section{Introduction}
\label{sec:introduction}

There have been recent advances
in computer vision and graphics,
enabling photo-realistic images to be created.
However, it still requires considerable skill or effort
to create a pixel-level detailed image from scratch.
Deep generative models,
such as generative adversarial networks (GANs) \cite{IGoodfellowNIPS2014}
and variational autoencoders (VAEs) \cite{DKingmaICLR2014,DRezendeICML2014},
have recently emerged as powerful models to alleviate this difficulty.
Although these models make it possible
to generate various images with high fidelity quality
by changing (e.g., randomly sampling)
latent variables in the generator or decoder input,
there still remains a painstaking process to create the desired image
because
the naive formulation does not impose any structure on latent variables;
as a result, they may be used by the generator or decoder
in a highly entangled manner.
This causes difficulty in interpreting
the ``meaning'' of the individual variables
and in controlling image generation by operating each one.

When we create an image from scratch,
we typically select and narrow a target to paint in a coarse-to-fine manner.
For example, when we create an image of a face with glasses,
we first roughly consider the type of glasses,
e.g., transparent glasses/sunglasses,
then define the details,
e.g., thin/thick rimmed glasses or small/big sunglasses.
To use a deep generative model
as a supporter for creating an image,
we believe that such {\it hierarchically interpretable representation}
is the key to obtaining the image one has in mind.

\begin{figure}[t]
\begin{center}
  \includegraphics[width=1.0\linewidth]{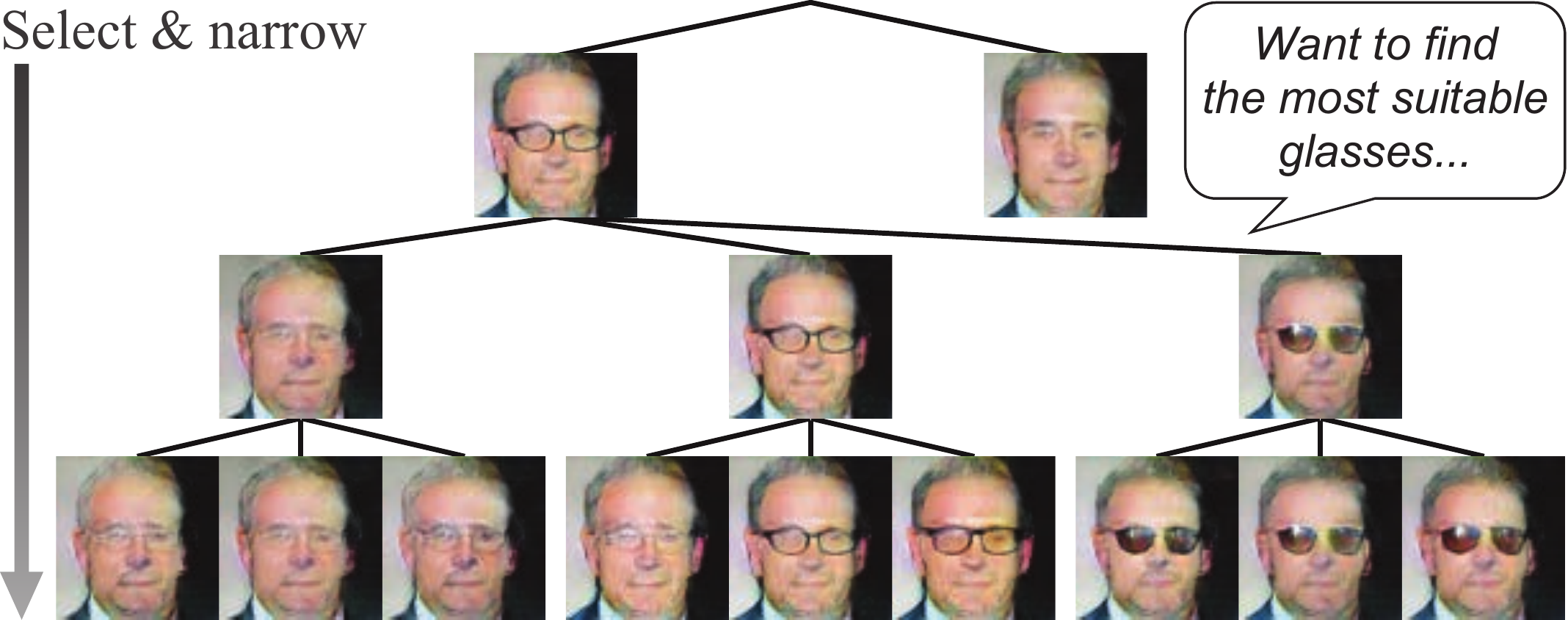}
\end{center}
\vspace{-2mm}
\caption{{\bf Examples of image generation
    under control using DTLC-GAN:}
    DTLC-GAN enables image generation to be controlled
    in coarse-to-fine manner, i.e., ``selected \& narrowed.''
    Our goal is to discover such hierarchically interpretable representations
    without relying on detailed supervision.}
\label{fig:concept}
\end{figure}

\begin{figure*}[t]
\begin{center}
  \includegraphics[width=1.0\textwidth]{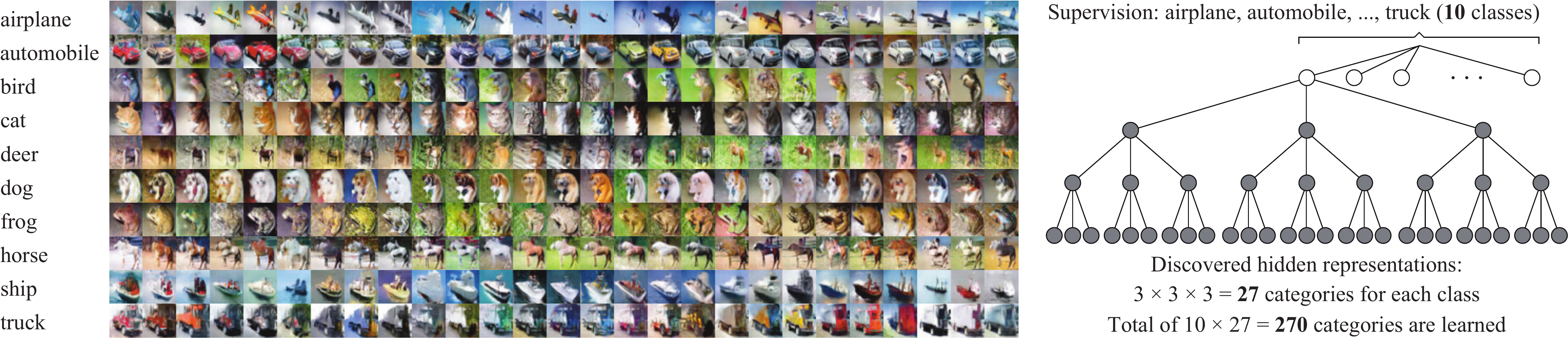}
\end{center}
\vspace{-2mm}
\caption{{\bf Generated image samples on CIFAR-10:}
  All images were generated from same noise but different latent codes.
  In each row, we varied second-, third-, and fourth-layer codes
  per nine images, per three images, and per image, respectively.
  Note that we learn these $10 \times 3 \times 3 \times 3 = 270$
  hierarchically disentangled representations
  only with supervision of class labels.
  This model also achieves high inception score {\bf 8.80}.
  We give details in Section~\ref{subsec:wgan-gp}.}
\label{fig:cifar10_all}
\end{figure*}

These facts motivated us to address the problem of
{\it how to derive hierarchically interpretable representations
in a deep generative model}.
To solve this problem,
we propose the decision tree latent controller GAN (DTLC-GAN),
an extension of the GAN that can
learn hierarchically interpretable representations
without relying on detailed supervision.
Figure~\ref{fig:concept} shows examples of image generation
under control using the DTLC-GAN.
If semantic features are represented in a hierarchically disentangled manner,
we can approach a target image gradually and interactively.

To impose a hierarchical inclusion structure
on latent variables,
we incorporate a new architecture called the DTLC into the generator input.
The DTLC has a multiple-layer tree structure in which
the ON or OFF of the child node codes is controlled
by the parent node codes.
By using this architecture hierarchically,
we can obtain the latent space
in which the lower layer codes are selectively used
depending on the higher layer codes.

On the problem of making the latent codes capture salient semantic features
of images in a hierarchically disentanglede manner in the DTLC,
the main difficulty is that
we need to discover representations disentangled
in the following three stages:
(1) disentanglement between the control target (e.g., glasses)
and unrelated factors (e.g., identity);
(2) coarse-to-fine disentanglement between layers,
i.e., the higher layer codes capture rough categories, while
the lower layer ones capture detailed categories;
and (3) inner-layer disentanglement
to control semantic features independently,
i.e., when one code captures a semantic feature
(e.g., thin glasses),
another one captures a different semantic feature
(e.g., thick glasses).

A possible solution would be to collect
detailed annotations,
the amount of which is large enough
to solve the problems in a fully supervised manner.
However, this approach incurs high annotation costs.
Even though we have enough human resources,
defining the detailed categories
remains a nontrivial task.
The latter problem is also addressed
in the field of research concerned with attribute representations
\cite{DParikhICCV2011,AYuICCV2015}
and is still an open issue.
This motivated us to 
tackle a challenging condition in which
hierarchically interpretable representations need to be learned
without relying on detailed annotations.
Under this condition,
it is not trivial to solve all the above three disentanglement problems
at the same time
because they are not independent from each other but are interrelated.
To mitigate these problems,
we propose a hierarchical conditional mutual information regularization (HCMI),
which is an extension of MI~\cite{XChenNIPS2016}
and conditional MI (CMI)~\cite{TKanekoCVPR2017}
to hierarchical conditional settings
and
optimize it with a newly defined curriculum learning \cite{YBengioICML2009}
method that we also propose.
This makes it possible to discover hierarchically interpretable representations
in a layer-by-layer manner
on the basis of information gain by only using a single DTLC-GAN model.
This is noteworthy because
we can learn expressive representations
without large increase in calculation cost.
Figure~\ref{fig:cifar10_all} shows typical examples on CIFAR-10, where
we succeeded in learning expressive representations,
i.e., $10 \times 3 \times 3 \times 3 = 270$ categories, are learned
in a weakly supervised
(i.e., only 10 class labels are supervised) manner.
We evaluated our DTLC-GAN on various datasets, i.e.,
MNIST, CIFAR-10, Tiny ImageNet, 3D Faces, and CelebA,
and confirmed that it can learn hierarchically interpretable representations
with either unsupervised or weakly supervised settings.
Furthermore, we applied our DTLC-GAN to image-retrieval tasks
and showed its effectiveness in representation learning.

\paragraph{Contributions:}
Our contributions are summarized as follows.
(1) We derive a novel functionality in a deep generative model,
which enables semantic features of an image to be controlled
in a coarse-to-fine manner.
(2) To obtain this functionality,
we incorporate a new architecture called the DTLC into a GAN,
which imposes a hierarchical inclusion structure on latent variables.
(3) We propose a regularization called the HCMI
and optimize it with a newly defined curriculum learning method
that we also propose.
This makes it possible to learn hierarchically disentangled representations
only using a single DTLC-GAN model without relying on detailed supervision.
(4) We evaluated our DTLC-GAN on various datasets
and confirmed its effectiveness in image generation and image-retrieval tasks.
We provide supplementary materials including demo videos at \url{http://www.kecl.ntt.co.jp/people/kaneko.takuhiro/projects/dtlc-gan/}.
  
\section{Related Work}
\label{sec:related}

\paragraph{Deep Generative Models:}
In computer vision and machine learning,
generative image modeling is a fundamental problem.
Recently, there was a significant breakthrough due to the emergence of
deep generative models.
These models roughly fall into two approaches:
deterministic and stochastic.
On the basis of deterministic approaches,
Dosovitsky et al.~\cite{ADosovitskiyCVPR2015}
proposed a deconvolution network that generates 3D objects,
and Reed et al.~\cite{SReedNIPS2015} and Yang et al.~\cite{JYangNIPS2015}
proposed networks that approximate functions for image synthesis.
There are three major models based on stochastic approaches.
One is a VAE
\cite{DKingmaICLR2014,DRezendeICML2014},
which is formulated as probabilistic graphical models
and optimized by maximizing a variational lower bound on the data likelihood.
Another is an autoregressive model~\cite{AOordICML2016},
which breaks the data distribution into a series of conditional distributions
and uses neural networks to model them.
The other is a GAN \cite{IGoodfellowNIPS2014},
which is composed of generator and discriminator networks.
The generator is optimized to fool the discriminator,
while the discriminator is optimized
to distinguish between real and generated data.
This min-max optimization makes the training procedure unstable,
but several techniques
\cite{MArjovskyICLR2017,MArjovskyICML2017,IGulrajaniNIPS2017,XMaoICCV2017,ARadfordICLR2016,TSalimansNIPS2016,JZhaoICLR2017}
have recently been proposed to stabilize it.
All these models have pros and cons.
In this paper, we take a stochastic approach, particularly focusing on a GAN,
and propose an extension to it
because it has flexibility on latent variable design.
Extension to other models remains a promising area for future work.

\paragraph{Disentangled Representation Learning:}
In the study of stochastic deep generative models,
there have been attempts to learn
disentangled representations similar to our attempt.
Most of the studies addressed the problem in supervised settings
and incorporated supervision into the networks.
For example, attribute/class labels
\cite{MMirzaArXiv2014,AOdenaICML2017,LTranCVPR2017,AOordNIPS2016,XYanECCV2016,ZZhangCVPR2017},
text descriptions
\cite{EMansimovICLR2016,SReedICML2016,HZhangICCV2017},
and object location descriptions
\cite{SReedNIPS2016,SReedICLRW2017}
are used as supervision.
To reduce the annotation cost,
extensions to semi-supervised settings have also recently been proposed
\cite{DKingmaNIPS2014,TSalimansNIPS2016,JSpringenbergICLR2016}.
The advantage of these settings is that
disentangled representation can be explicitly learned following the supervision;
however,
the limitation is that 
learnable representations are restricted to supervision.
To overcome this limitation,
weakly supervised
\cite{TKanekoCVPR2017,TKulkarniNIPS2015,AMakhzaniNIPS2016,MMathieuNIPS2016}
and unsupervised \cite{XChenNIPS2016} models
have recently been proposed,
which discover meaningful hidden representation
without relying on detailed annotations;
however, these models are limited to discovering
one-layer hidden representations,
whereas the DTLC-GAN enables multi-layer hidden representations to be learned.
We further discussed the relationship to the previous GANs
in Section~\ref{subsec:relation}.

\paragraph{Hierarchical Representation:}
The other related topic is hierarchical representation.
Previous studies have decomposed an image in various ways.
The LAPGAN \cite{EDentonNIPS2015} and StackGAN \cite{HZhangICCV2017}
deconstruct an image by repeatedly downsampling it,
S$^2$-GAN \cite{XWangECCV2016}
decomposes the generative process to structure and style,
VGAN \cite{CVondrickNIPS2016} decomposes
a video into foreground and background, and
SGAN~\cite{XHuangCVPR2017} learns multi-level representations
in feature spaces of intermediate layers.
Other studies \cite{SEslamiNIPS2016,KGregorICML2015,DImICLRW2016,HKwakArXiv2016,JYangICLR2017}
used recursive structures to draw images in a step-by-step manner.
The main difference from these studies is
that they derive hierarchical representations
in a pixel space or feature space
to improve the fidelity of an image,
while we derive those in a latent space
to improve the interpretability and controllability of latent codes.
More recently, Zhao et al.~\cite{SZhaoICML2017} proposed
an extension of a VAE called the VLAE
to learn multi-layer hierarchical representations in a latent space
similar to ours;
however, the type of hierarchy is different from ours.
They learn representations that are
semantically independent among layers, whereas
we learn those
where lower layer codes are correlated with higher layer codes
in a decision tree manner.
We argue that such representation is necessary to learn
category-specific features
and control image generation in a select-and-narrow manner.

\section{Background: GAN}
\label{sec:gan}

A GAN \cite{IGoodfellowNIPS2014} is a framework
for training a generative model using a min-max game.
The goal is to learn generative distribution $P_{G}({\bm x})$
that matches the real data distribution $P_{\rm data}({\bm x})$.
It consists of two networks:
a generator $G$
that transforms noise ${\bm z} \sim P_{\bm z}({\bm z})$ into
data space ${\bm x} = G({\bm z})$,
and a discriminator $D$ that assigns probability
$p = D({\bm x}) \in [0, 1]$
when ${\bm x}$ is a sample from $P_{\rm data}$
and assigns probability $1 - p$
when it is a sample from $P_{G}$.
The $P_{\bm z}({\bm z})$ is a prior on ${\bm z}$.
The $D$ and $G$ play a two-player min-max game with
the following binary cross entropy:
\begin{align}
  \label{eqn:gan}
  {\cal L}_{\rm GAN}(D, G) =
  & \: \mathbb{E}_{{\bm x} \sim P_{\rm data}({\bm x})}[ \log D({\bm x}) ]
  \nonumber \\
  +
  & \: \mathbb{E}_{{\bm z} \sim P_{\bm z}({\bm z})}
  [ \log ( 1 - D(G({\bm z})) ) ].
\end{align}
The $D$ attempts to find the binary classifier
for discriminating between
true and generated data by maximizing this loss, whereas
the $G$ attempts to generate data indistinguishable from the true data
by minimizing this loss.

\begin{figure*}[t]
\begin{center}
  \includegraphics[width=1.0\linewidth]{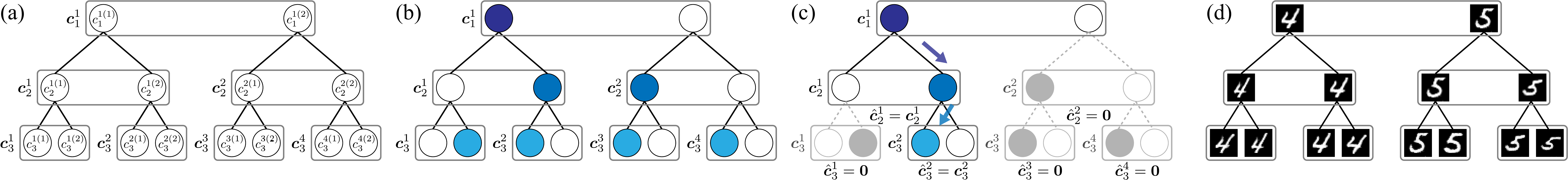}
\end{center}
\vspace{-2mm}
\caption{{\bf Sampling example using three-layer DTLC:}
  (a) Architecture of three-layer DTLC where $k_1, k_2, k_3 = 2$.
  (b) Sampling example in Step 1.
  Each code is sampled from categorical distribution.
  Filled and open circles indicate 1 and 0, respectively.
  (c) Sampling example in Steps 2 and 3.
  ON or OFF of child node codes is selected by parent node codes.
  This execution is conducted recursively from highest layer to lowest layer.
  This imposes hierarchical inclusion constraints on sampling.
  (d) Sample images generated using this controller.
  Each image corresponds to each latent code.
  We tested on subset of MNIST dataset,
  which includes ``4'' and ``5'' digit images.
  This is relatively easy dataset;
  however, it is noteworthy that
  hierarchically disentangled representations,
  such as ``4'' or ``5'' in first layer and
  ``narrow-width 4'' or ``wide-width 4'' in second left layer, are learned
  in fully unsupervised manner.
}
\label{fig:dtlc}
\end{figure*}

\section{DTLC-GAN}
\label{sec:dtlc-gan}

\subsection{DTLC}
\label{subsec:dtlc}
In the naive GAN,
latent variables
are sampled from an unconditional prior
and do not have any constraints on a structure.
As a result, they may be used by the $G$ in a highly entangled manner,
causing difficulty in interpreting
the ``meaning'' of the individual variables
and in controlling image generation by operating each one.
Motivated by this fact, we incorporate the DTLC into the generator input
to impose a hierarchical inclusion structure on latent variables.

\paragraph{Notation:}
In the DTLC-GAN, the latent variables are decomposed into multiple levels.
We first decompose the latent variables into two parts:
$\hat{\bm c}_L$, which is a latent code derived from an $L$-layer DTLC and
will target hierarchically interpretable semantic features, and
${\bm z}$, which is a source of incompressible noise 
that covers factors that are not represented by $\hat{\bm c}_L$.
To derive $\hat{\bm c}_L$,
the DTLC has a multiple-layer tree structure and is composed of $L$ layer codes
${\bm c}_1, \cdots, {\bm c}_L$.
In each layer, ${\bm c}_l$ is decomposed into $N_l$ node codes
${\bm c}_l = ( {\bm c}_l^1, \cdots, {\bm c}_l^{N_l} )$.
To impose a hierarchical inclusion relationship
between the $l$th and $(l+1)$th layers,
an $n$th parent node code ${\bm c}_{l}^n$ is associated with
$k_{l}$ child node codes
${\bm c}_{l+1}^{n, 1}, \cdots, {\bm c}_{l+1}^{n, k_{l}}$,
where ${\bm c}_{l+1}^{n, i} = {\bm c}_{l+1}^{k_l (n - 1) + i}$.
By this definition,
$N_{l+1}$ is calculated as $N_{l+1} = N_l \times k_l$.

We can use both discrete and continuous variables as
${\bm c}_l^n$, but
for simplicity, we treat the case in which
parent node codes are discrete and
the lowest layer codes are either discrete or continuous.
In this case,
${\bm c}_l^n$
is represented as a $k_l$ dimensional onehot vector
and each dimension $c_l^{n (i)}$ $(i = 1, \cdots, k_{l})$
is associated with one child node code ${\bm c}_{l+1}^{n, i}$.

\paragraph{Sampling Scheme:}
In a training phase, we sample latent codes as follows.
We illustrate a sampling example in Figure~\ref{fig:dtlc}.

\begin{enumerate}
\item We sample ${\bm c}_l^n$ $(l = 1, \cdots, L-1)$
  from categorical distribution
  ${\bm c}_l^n \sim {\rm Cat} \left( K = k_{l}, p = \frac{1}{k_{l}} \right)$.
  We sample ${\bm c}_L^n$ in a similar manner in the discrete case, while
  we sample it from uniform distribution
  ${\bm c}_L^n \sim {\rm Unif}(-1, 1)$ in the continuous case.
  Note that, if we have supervision for ${\bm c}_l^n$,
  we can directly use it instead of sampling.

\item To impose a hierarchical inclusion structure,
  we sample $\hat{\bm c}_{l+1}^{n, i}$ from conditional prior
  $\hat{\bm c}_{l+1}^{n, i} \sim P ( \hat{\bm c}_{l+1}^{n, i} | \hat{\bm c}_l^n )$, where $\hat{\bm c}_1 = {\bm c}_1$.
  We do this with the following process:\\
  \begin{equation}
    \hat{\bm c}_{l+1}^{n, i} = \hat{c}_l^{n (i)} {\bm c}_{l+1}^{n, i},
  \end{equation}
  where $\hat{c}_l^{n(i)}$ is the $i$th dimension of $\hat{\bm c}_l^n$.
  This equation means that a parent node code acts as a child node selector
  controlling the ON or OFF of a child node code.

\item By executing Step 2 recursively
  from the highest layer to the lowest layer, we can sample
  $\hat{\bm c}_L$
  with $L$ layer hierarchical inclusion constraints.
  We add it to the generator input and
  use it with ${\bm z}$ to generate an image:
  ${\bm x} = G(\hat{\bm c}_L, {\bm z})$.
\end{enumerate}

\begin{figure*}[t]
\begin{center}
  \includegraphics[width=1.0\linewidth]{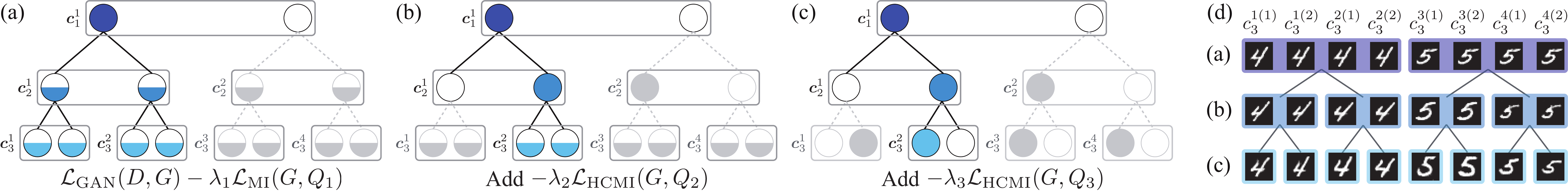}
\end{center}
\vspace{-2mm}
\caption{{\bf Example of curriculum learning:}
  (a) We first learn disentangled representations in first layer.
  To do this, we only use regularization for this layer and
  fix and set average value to lower layer codes.
  (b)(c) We then learn disentangled representations in second and third layers
  in layer-by-layer manner.
  We add regularization and sampling in turn depending on training phase.
  (d) Image samples generated in each phase.
  In phase (a), first-layer codes are learned,
  while second- and third-layer codes are fixed; therefore,
  $2$ disentangled representations are obtained.
  In phase (b), first- and second-layer codes are learned,
  while third-layer codes are fixed; therefore,
  $2 \times 2$ disentangled representations are obtained.
  In phase (c), all codes are learned; therefore,
  $2 \times 2 \times 2$ disentangled representations are obtained.
}
\label{fig:curriculum}
\end{figure*}

\subsection{HCMI}
\label{subsec:regularization}
The DTLC imposes a hierarchical inclusion structure on latent variables;
however, its constraints are not
sufficient to correlate latent variables with
semantic features of images.
To solve this problem without relying on detailed supervision,
we propose a hierarchical conditional mutual information regularization (HCMI),
which is an extension of MI~\cite{XChenNIPS2016}
and conditional MI (CMI)~\cite{TKanekoCVPR2017}
to hierarchical conditional settings.
In particular, we use different types of regularization
for the second layer to the $L$th layer, which have parent node codes,
and the first layer, which does not have those.

\paragraph{Regularization for Second Layer to $L$th Layer:}
In this case, we need
to discover semantic features in a hierarchically restricted manner;
therefore, we maximize mutual information between
$l$th-layer child node code
${\bm c}$
and
image $G(\hat{\bm c}_L, {\bm z})$
conditioned on $(l-1)$th-layer parent node code
${\bm p}$:
$I({\bm c}; G(\hat{\bm c}_L, {\bm z}) | {\bm p})$.
For simplicity, we denote
$\hat{\bm c}_{l}^{n, i}$ and $\hat{\bm c}_{l-1}^n$
as ${\bm c}$ and ${\bm p}$, respectively.
In practice, exact calculation of this mutual information is difficult
because it requires calculation of the intractable posterior
$P({\bm c} | {\bm x}, {\bm p})$.
Therefore,
following previous studies~\cite{XChenNIPS2016,TKanekoCVPR2017},
we instead calculate its lower bound using
an auxiliary distribution
$Q({\bm c} | {\bm x}, {\bm p})$
approximating
$P({\bm c} | {\bm x}, {\bm p})$:
\begin{flalign}
  \label{eqn:hcmi}
  & \: I({\bm c}; G(\hat{\bm c}_L, {\bm z}) | {\bm p})
  \nonumber \\
  = & \: H({\bm c} | {\bm p})
  - H({\bm c} | G(\hat{\bm c}_L, {\bm z}), {\bm p})
  \nonumber \\
  = & \: H({\bm c} | {\bm p})
  + \mathbb{E}_{{\bm x} \sim G(\hat{\bm c}_L, {\bm z})}[
    \mathbb{E}_{{\bm c}' \sim P({\bm c}|{\bm x}, {\bm p})}[
      \log P({\bm c}' | {\bm x}, {\bm p})]]
  \nonumber \\
  = & \: H({\bm c} | {\bm p})
  + \mathbb{E}_{{\bm x} \sim G(\hat{\bm c}_L, {\bm z})}[
    D_{\rm KL}(P(\cdot | {\bm x}, {\bm p}) || Q(\cdot | {\bm x}, {\bm p}))
    \nonumber \\
    & \: + \mathbb{E}_{{\bm c}' \sim P({\bm c}|{\bm x}, {\bm p})}[
      \log Q({\bm c}' | {\bm x}, {\bm p})]]
  \nonumber \\
  \geq & \: H({\bm c} | {\bm p})
  + \mathbb{E}_{{\bm x} \sim G(\hat{\bm c}_L, {\bm z})}[
    \mathbb{E}_{{\bm c}' \sim P({\bm c}|{\bm x}, {\bm p})}[
      \log Q({\bm c}' | {\bm x}, {\bm p})]]
  \nonumber \\
  = & \: H({\bm c} | {\bm p})
  + \mathbb{E}_{{\bm c} \sim P({\bm c} | {\bm p}),
    {\bm x} \sim G(\hat{\bm c}_L, {\bm z})}
  [\log Q({\bm c} | {\bm x}, {\bm p})].
\end{flalign}
For simplicity, we fix the distribution of ${\bm c}$ and
treat $H({\bm c} | {\bm p})$ as constant.
In practice, $Q$ is parametrized as a neural network and
we particularly denote the network for
$\hat{\bm c}_{l}^m$~$(= \hat{\bm c}_{l}^{n, i})$ as
$Q_{l}^m$, where $m = k_{l-1}(n - 1) + i$.
Thus, the final objective function is written as
\begin{flalign}
  \label{eqn:hcmi}
  & {{\cal L}_{\rm HCMI}}(G, Q_{l}^{m}) \nonumber \\
  = & \mathbb{E}_{{\bm c} \sim P({\bm c} | {\bm p}),
    {\bm x} \sim G(\hat{\bm c}_L, {\bm z})}
  [\log Q_{l}^m({\bm c} | {\bm x}, {\bm p})].
\end{flalign}
The $Q_{l}^m$ attempts to discover the specific semantic features
that correlate with ${\bm c}$
in terms of conditional information gain by maximizing this objective.
We calculate ${{\cal L}_{\rm HCMI}}(G, Q_{l}^{m})$ for every child node code
$\hat{\bm c}_{l}^m$.
We denote the summation in the $l$th layer as
${\cal L}_{\rm HCMI}(G, Q_{l}) = \sum_{m=1}^{N_{l}} {{\cal L}_{\rm HCMI}}(G, Q_{l}^{m})$.
We use this objective with trade-off parameter $\lambda_{l}$.

\paragraph{Regularization for First Layer:}
The above regularization is useful
for the codes that have parent node codes;
however, the first-layer codes do not have those; thus,
we instead use a different regularization for them.
Fortunately, this single-layer case has been addressed
in previous studies~\cite{XChenNIPS2016,AOdenaICML2017}
and we use one of them depending on the supervision setting.
In an unsupervised setting,
we use the MI \cite{XChenNIPS2016} written as
\begin{flalign}
  \label{eqn:mi}
  {\cal L}_{\rm MI}(G, Q_1)
  = \mathbb{E}_{{\bm c}_1 \sim P({\bm c}_1),
    {\bm x} \sim G(\hat{\bm c}_L, {\bm z})}
  [\log Q_1 ({\bm c}_1 | {\bm x})].
\end{flalign}
In a weakly supervised setting,
we use an auxiliary classifier regularization (AC) \cite{AOdenaICML2017}
written as
\begin{flalign}
  \label{eqn:ac}
  {\cal L}_{\rm AC}(G, Q_1)
  = & \: \mathbb{E}_{{\bm c}_1 \sim P({\bm c}_1),
    {\bm x} \sim G(\hat{\bm c}_L, {\bm z})}
  [\log Q_1 ({\bm c}_1 | {\bm x})]
  \nonumber \\
  + & \: \mathbb{E}_{{\bm c}_1, {\bm x} \sim P_{\rm data}({\bm c}_1, {\bm x})}
  [\log Q_1 ({\bm c}_1 | {\bm x})].
\end{flalign}
Note that the first term is the same as ${\cal L}_{\rm MI}(G, Q_1)$,
and the added second term acts as supervision regularization.
We use these objectives with trade-off parameter $\lambda_1$.

\paragraph{Full Objective:}
Our full objective is written as
\begin{flalign}
  \label{eqn:full}
  & \: {\cal L}_{\rm Full}(D, G, Q_1, \cdots, Q_L)
  \nonumber \\
  = & \: {\cal L}_{\rm GAN}(D, G)
  - \lambda_1 {\cal L}_{\rm MI/AC}(G, Q_1)
  - \sum_{l=2}^L \lambda_l {\cal L}_{\rm HCMI}(G, Q_l).
\end{flalign}
This is minimized for the $G$ and $Q_1, \cdots, Q_L$
and maximized for the $D$.

\subsection{Curriculum Learning}
\label{subsec:learning}
The HCMI works well when the higher layer codes are already known;
however,
we assume the condition in which
detailed annotations are not provided in advance.
As a result,
the network may confuse between
inner-layer and intra-layer disentanglement
at the beginning of training.
To mitigate this problem,
we developed our curriculum learning method.
In particular, we define a curriculum for regularization and sampling.
We illustrate an example of the proposed curriculum learning method
in Figure~\ref{fig:curriculum}.

\paragraph{Curriculum for Regularization:}
As a curriculum for regularization,
we do not use the whole regularization
in Equation~\ref{eqn:full} at the same time,
instead, we add the regularization from the highest layer to the lowest layer
in turn
according to the training phase.
In an unsupervised setting,
we first learn with
${\cal L}_{\rm GAN}(D, G) - \lambda_1 {\cal L}_{\rm MI}(G, Q_1)$
then add $-\lambda_2 {\cal L}_{\rm HCMI}(G, Q_2), \cdots, -\lambda_L {\cal L}_{\rm HCMI}(G, Q_L)$ in turn.
In a weakly supervised setting,
we first learn with
${\cal L}_{\rm GAN}(D, G) - \lambda_1 {\cal L}_{\rm AC}(G, Q_1) - \lambda_2 {\cal L}_{\rm HCMI}(G, Q_2)$
then add $-\lambda_3 {\cal L}_{\rm HCMI}(G, Q_3), \cdots, -\lambda_L {\cal L}_{\rm HCMI}(G, Q_L)$ in turn.
We use different curricula between these two settings
because in a weakly supervised setting,
we already know the first-layer codes; thus,
we can start from learning the second-layer codes.

\paragraph{Curriculum for Sampling:}
In learning the higher layer codes,
instability caused by random sampling of the lower layer codes
can degrade the learning performance.
Motivated by this fact, we define a curriculum for sampling.
In particular,
in learning the higher layer codes $\hat{\bm c}_l$,
we fix and set the average value to the lower layer codes
${\bm c}_{l+1}, \cdots, {\bm c}_L$,
e.g., set $\frac{1}{k_{l+1}}$ for discrete code ${c}_{l+1}^{n(i)}$
and set $0$ for continuous code ${c}_{l+1}^{n(i)}$.

\subsection{Relationship to Previous GANs}
\label{subsec:relation}
The DTLC-GAN is a general framework, and
we can see it as a natural extension of previous GANs.
We summarize this relationship in Table~\ref{tab:relation}.
In particular, the InfoGAN~\cite{XChenNIPS2016} and
CFGAN~\cite{TKanekoCVPR2017}\footnote{Strictly speaking, the CFGAN is formulated
  as an extension of the CGAN,
  while the weakly supervised DTLC-GAN
  is formulated as an extension of the AC-GAN.
  Therefore, these two models do not have completely the same architecture;
  however, they share the similar motivation.}
are highly related to the DTLC-GAN
in terms of discovering hidden representations on the basis of information gain;
however, they are limited to learning one-layer hidden representation.
We developed our DTLC-GAN,
HCMI, and curriculum learning method to overcome this limitation.

\begin{table}
  \begin{center}
    \footnotesize{
      \begin{tabular}{l|c|c}
        {\bf \# of Hidden}
        & {\bf Unsupervised} & {\bf (Weakly) Supervised} \\
        {\bf Layers} & &
        \\ \Xhline{0.8pt}
        0 & GAN \cite{IGoodfellowNIPS2014}
        & CGAN \cite{MMirzaArXiv2014}\footnotemark[1],
        AC-GAN \cite{AOdenaICML2017} \\
        1 & InfoGAN \cite{XChenNIPS2016}
        & CFGAN \cite{TKanekoCVPR2017}\footnotemark[1] \\
        \cline{2-3}
        2 & \multicolumn{2}{c}{} \\
        3 & \multicolumn{2}{c}{\bf DTLC-GAN} \\
        4, $\cdots$ & \multicolumn{2}{c}{} \\
      \end{tabular}
    }
  \end{center}
  \vspace{-2mm}
  \caption{Relationship to previous GANs}
  \label{tab:relation}
\end{table}

\section{Implementation}
\label{sec:implementation}
We designed the network architectures and training scheme
on the basis of techniques
introduced for the InfoGAN~\cite{XChenNIPS2016}.
The $D$ and $Q_1, \cdots, Q_L$ share all convolutional layers, and
one fully connected layer is added to the final layer for $Q_l$.
This means that the difference in the calculation cost for the GAN and DTLC-GAN
is negligibly small.
For discrete code $\hat{\bm c}_l^m$,
we represent $Q_l^m$
as softmax nonlinearity.
For continuous code $\hat{\bm c}_l^m$,
we treat $Q_l^m$
as a factored Gaussian.

In most of the experiments we conducted,
we used typical DCGAN models~\cite{ARadfordICLR2016}
and did not use the state-of-the-art GAN training techniques
to evaluate whether the DTLC-GAN works well without relying on such techniques.
However, our contributions are orthogonal to these techniques;
therefore, we can improve image quality easily by incorporating
these techniques to our DTLC-GAN.
To demonstrate this, we also tested the DTLC-WGAN-GP
(our DTLC-GAN with the WGAN-GP ResNet~\cite{IGulrajaniNIPS2017})
as discussed in Section~\ref{subsec:wgan-gp}.
The details of the experimental setup are
given in Section~\ref{sec:detail} of the appendix.

\section{Experiments}
\label{sec:experiments}
We conducted experiments on various datasets,
i.e., MNIST \cite{YLeCunIEEE1998},
CIFAR-10 \cite{AKrizhevskyTech2009},
Tiny ImageNet~\cite{ORussakovskyIJCV2015},
3D Faces \cite{PPaysanAVSS2009}, and
CelebA \cite{ZLiuICCV2015},
to evaluate the effectiveness and generality of
our DTLC-GAN.\footnote{Due to the limited space,
  we provide only the important results in this main text.
  Please refer to the appendix for more results.}
We first used the MNIST and CIFAR-10 datasets,
which are widely used in this field,
to analyze the DTLC-GAN qualitatively and quantitatively.
In particular, we evaluated the DTLC-GAN in an unsupervised setting
on the MNIST dataset
and
in a weakly supervised setting on the CIFAR-10 dataset
(Section~\ref{subsec:unsupervised} and \ref{subsec:weaklysupervised},
respectively).
We tested the DTLC-WGAN-GP on the CIFAR-10 and Tiny ImageNet datasets
to demonstrate that our contributions are orthogonal to the state-of-the-art
GAN training techniques (Section~\ref{subsec:wgan-gp}).
We used the 3D Faces dataset
to evaluate the effectiveness of the DTLC-GAN with continuous codes
(Section~\ref{subsec:continuous}) and
evaluated it on image-retrieval tasks using the CelebA dataset
(Section~\ref{subsec:retrieval}).
Hereafter, we denote the DTLC-GAN with an $L$th layer DTLC
as the {\bf DTLC$^L$-GAN} and DTLC$^L$-GAN in a weakly supervised setting
as the {\bf DTLC$^L$-GAN$_{\rm WS}$}.

\begin{figure}[t]
\begin{center}
  \includegraphics[width=1.0\linewidth]{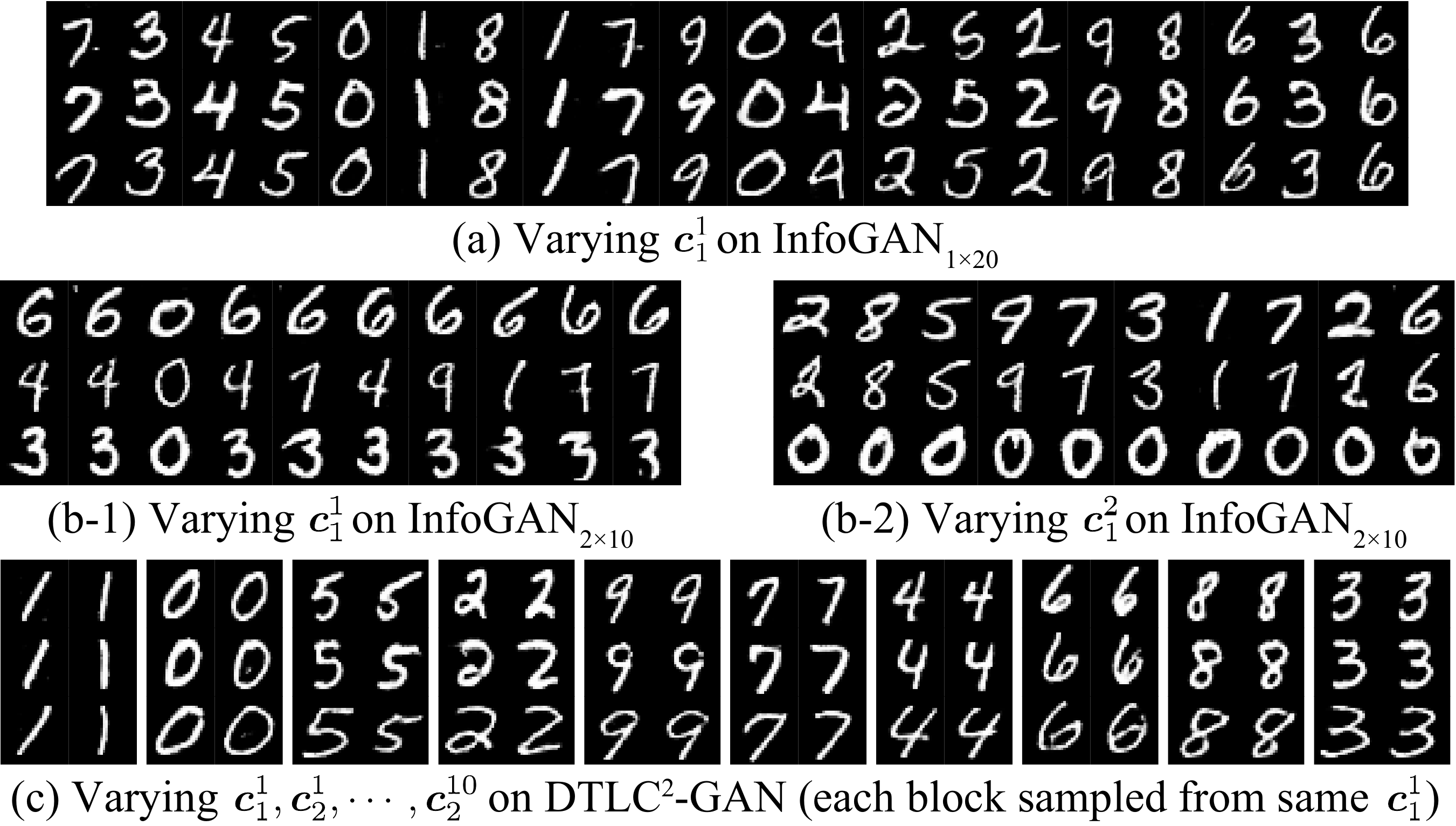}
\end{center}
\vspace{-2mm}
\caption{{\bf Representation comparison on MNIST:}
  We compared models
  in which dimensions of latent codes given to $G$ are same 20.
  In each figure,
  column contains three samples from same category.
  In each row, one latent code is varied,
  while other latent codes and noise are fixed.}
\label{fig:representation_comparison}
\end{figure}

\subsection{Unsupervised Representation Learning}
\label{subsec:unsupervised}
We first analyzed the DTLC-GAN in unsupervised settings on
the MNIST dataset,
which consists of photographs of handwritten digits and contains
60,000 training and 10,000 test samples.

\paragraph{Representation Comparison:}
To confirm the effectiveness of the hierarchical representation learning,
we compared the DTLC-GAN with that
in which dimensions of latent codes given to the $G$ are the same
but are not hierarchical.
To represent our DTLC-GAN,
we used the {\bf DTLC$^2$-GAN},
where $k_1 = 10$ and $k_2 = 2$.
In this model, $\hat{\bm c}_2$, the dimension of which is
$10 \times 2 = 20$, is given to the $G$.
For comparison, we used two models in which latent code dimensions are also 20
but not hierarchical.
One is the {\bf InfoGAN$_{1 \times 20}$}, which has one code
${\bm c}_1^1 \sim {\rm Cat}(K=20, p=0.05)$, and
the other is the {\bf InfoGAN$_{2 \times 10}$}, which has two codes
${\bm c}_1^1, {\bm c}_1^2 \sim {\rm Cat}(K=10, p=0.1)$.
We show the results in Figure~\ref{fig:representation_comparison}.
In (c), the {\bf DTLC$^2$-GAN} succeeded in
learning hierarchically interpretable representations
(in the first layer, digits, and in the second layer, details of each digit).
In (a), the {\bf InfoGAN$_{1 \times 20}$} succeeded in learning
disentangled representations; however,
they were learned as a flat relationship; thus,
it was not trivial to estimate
the higher concept (e.g., digits) from them.
In (b-1) and (b-2),
the {\bf InfoGAN$_{2 \times 10}$} failed to learn interpretable representations.
We argue that this is because
${\bm c}_1^1$ and ${\bm c}_1^2$ struggle to represent digit types.
To clarify this limitation, we also conducted experiments
on simulated data.
See Section~\ref{subsec:simulation} of the appendix for details.

\begin{figure}[t]
\begin{center}
  \includegraphics[width=0.945\linewidth]{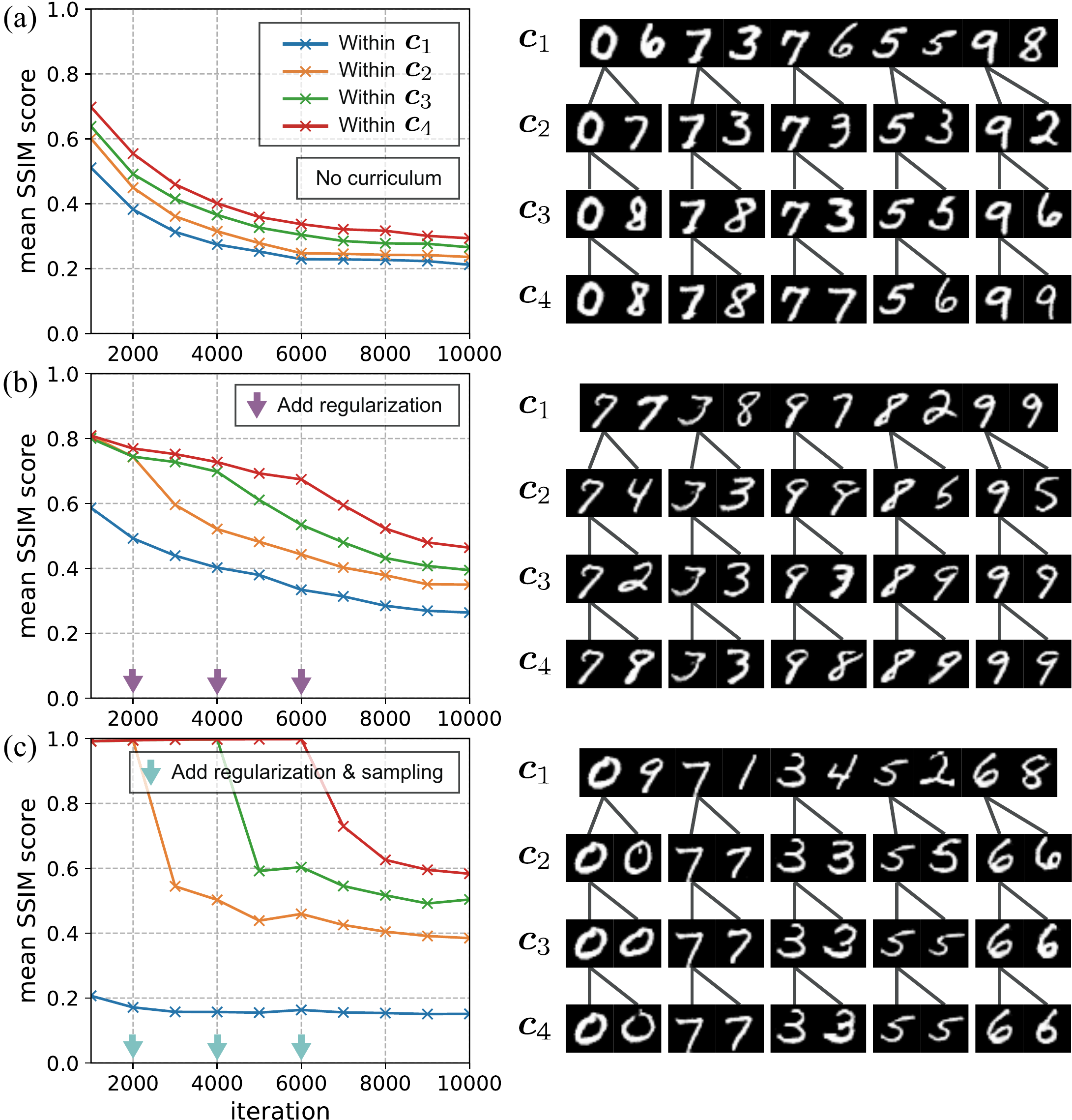}
\end{center}
\vspace{-2mm}
\caption{{\bf Ablation study in unsupervised settings on MNIST:}
  Left figures show changes in mean SSIM scores through learning.
  We measured those between pairs of images within same category per layer.
  Right figures show sample images generated
  with varying latent codes per layer.
  Gray line indicates parent-child relationship.
  From top to bottom,
  (a) DTLC$^4$-GAN without curriculum,
  (b) DTLC$^4$-GAN with curriculum for regularization, and
  (c) DTLC$^4$-GAN with full curriculum
  (curriculum for regularization and sampling:
  proposed curriculum learning method).}
\vspace{-3mm}
\label{fig:mnist_cd}
\end{figure}

\begin{figure}[t]
\begin{center}
  \includegraphics[width=1.0\linewidth]{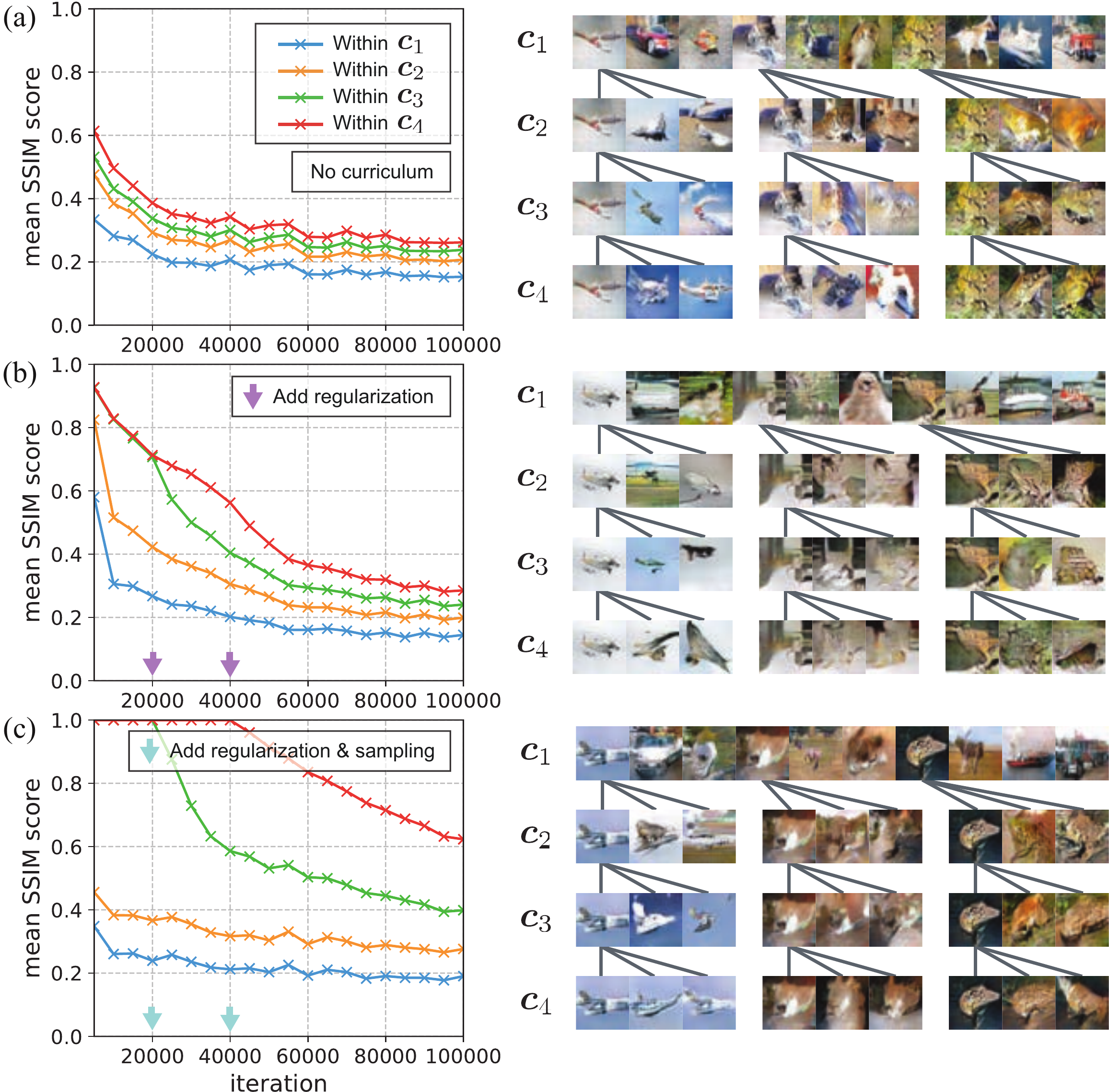}
\end{center}
\vspace{-2mm}
\caption{{\bf Ablation study in weakly supervised settings on CIFAR-10:}
  View of figure is similar to that in Figure~\ref{fig:mnist_cd}}
\vspace{-2mm}
\label{fig:cifar10_cd}
\end{figure}

\paragraph{Ablation Study on Curriculum Learning:}
To analyze the effectiveness of the proposed curriculum learning method,
we conducted an ablation study.
To evaluate quantitatively,
we measured the inter-category diversity of generated images
on the basis of structural similarity (SSIM) \cite{ZWangIP2004},
which is a well-characterized perceptual similarity metric.
This is an ad-hoc measure; however,
recent studies \cite{TKanekoCVPR2017,AOdenaICML2017}
showed that an SSIM-based measure is useful for evaluating
the diversity of images generated with a GAN.
Note that
evaluating the quality of deep generative models is not trivial
and is still an open issue
due to the variety of probabilistic criteria~\cite{LTheisICLR2016}.
To evaluate the $l$th layer inter-category diversity,
we measured the SSIM scores between pairs of images
that are sampled from the same noise and higher layer codes
but random $l$th and lower layer codes.
We calculated the scores for 50,000 randomly sampled pairs of images and
took the average.
The smaller value indicates that diversity is larger.
We show changes in the mean SSIM scores through learning
and sample images generated with varying latent codes per layer
in Figure~\ref{fig:mnist_cd}.
We used the {\bf DTLC$^4$-GAN}, where $k_1=10$ and $k_2, k_3, k_4 = 2$.
From these results, the {\bf DTLC$^4$-GAN} with the full curriculum succeeded in
making higher layer codes obtain higher diversity
and lower layer codes obtain lower diversity,
while the others failed.
We argue that this is because the latter
cannot avoid confusion between inner-layer and intra-layer disentanglement.
The qualitative results also support this fact.
We also show sample images for all categories in
Figures~\ref{fig:mnist_all_l0_s0_ex}--\ref{fig:mnist_all_ex} of the appendix.

\subsection{Weakly Supervised Representation Learning}
\label{subsec:weaklysupervised}
We next analyzed the DTLC-GAN in weakly supervised settings
(i.e., only class labels are supervised)
on the CIFAR-10 dataset,
which consists of 10 classes of images and
contains 5,000 training and 1,000 test samples per class.

\paragraph{Ablation Study on Curriculum Learning:}
We conducted an ablation study
to evaluate the effectiveness of the proposed curriculum learning method
in weakly supervised settings.
We show changes in mean SSIM scores through learning
and sample images generated with varying latent codes per layer
in Figure~\ref{fig:cifar10_cd}.
In this experiment,
we used the {\bf DTLC$^4$-GAN$_{\rm WS}$},
where $k_1=10$ and $k_2, k_3, k_4 = 3$.
We can see the same tendency as in Figure~\ref{fig:mnist_cd}.
These results indicate that
proposed curriculum learning method is indispensable,
even in weakly supervised settings.
We show samples images for all categories in
Figures~\ref{fig:cifar10_all_l0_s0_ex}--\ref{fig:cifar10_all_ex}
of the appendix.
We also conducted preference tests
to analyze visual interpretability.
See Section~\ref{subsec:interpretability} of the appendix for details.

\paragraph{Quantitative Evaluation:}
An important concern is whether
our extension degrades image quality.
To address this concern,
we evaluated the DTLC-GAN$_{\rm WS}$ on three metrics:
inception score~\cite{TSalimansNIPS2016},
adversarial accuracy~\cite{JYangICLR2017}, and
adversarial divergence~\cite{JYangICLR2017}.\footnote{The
  latter two metrics require
  pairs of generated images and class labels to train a classifier.
  In our settings, a conditional generator is learned; thus,
  we directly used it to generate an image with a class label.
  We used the classifier, architecture of which was similar to the $D$
  except for the output layer.}
We list the results in Table~\ref{tab:quantitative}.
We compared a {\bf GAN}, the {\bf AC-GAN}, and {\bf DTLC$^L$-GAN$_{\rm WS}$},
where $k_1=10$ and $k_2, \cdots, k_L=3$.
For fair comparison, we used the same network architecture and training scheme
except for the extended parts.
The inception scores are not state-of-the-art,
but in this comparison,
the {\bf DTLC$^L$-GAN$_{\rm WS}$}
improved upon {\bf GAN} and was comparable to the {\bf AC-GAN}.
The adversarial accuracy and adversarial divergence scores are
state-of-the art, and
the {\bf DTLC$^L$-GAN$_{\rm WS}$} improved upon the {\bf AC-GAN}.
These results are noteworthy because they indicate that
we can obtain expressive representation using the DTLC-GAN$_{\rm WS}$
without concern for image-quality degradation.

\begin{table}
  \begin{center}
    \footnotesize{
      \begin{tabular*}{\columnwidth}{lccc}
        & {\bf Inception} & {\bf Adversarial} & {\bf Adversarial} \\
        {\bf Model} & {\bf Score}     & {\bf Accuracy}    & {\bf Divergence}
        \\ \Xhline{0.8pt}        
        GAN
        & 7.09 $\pm$ 0.09 & - & - \\
        AC-GAN
        & 7.41 $\pm$ 0.06 & 50.99 $\pm$ 0.55 & 2.07 $\pm$ 0.02 \\
        DTLC$^2$-GAN$_{\rm WS}$
        & 7.39 $\pm$ 0.03 & 55.10 $\pm$ 0.48 & {\bf 1.82 $\pm$ 0.03} \\
        DTLC$^3$-GAN$_{\rm WS}$
        & 7.35 $\pm$ 0.09  & 55.20 $\pm$ 0.47 & 1.95 $\pm$ 0.05 \\
        DTLC$^4$-GAN$_{\rm WS}$
        & 7.46 $\pm$ 0.06  & 56.19 $\pm$ 0.36 & 1.93 $\pm$ 0.05 \\
        DTLC$^5$-GAN$_{\rm WS}$
        & {\bf7.51 $\pm$ 0.06} & {\bf 58.87 $\pm$ 0.52} & 1.83 $\pm$ 0.04 \\
        \hline
        Real Images
        & 11.24 $\pm$ 0.12 & 85.77 $\pm$ 0.22 & 0 \\
        State-of-the-Art
        & 8.59 $\pm$ 0.12\dag\!\!\! &
        44.22 $\pm$ 0.08\ddag\!\!\! &
        5.57 $\pm$ 0.06\ddag\!\!\! \\
      \end{tabular*}
    }
  \end{center}
  \vspace{-2mm}
  \caption{Quantitative comparison between GAN, AC-GAN, and DTLC-GAN$_{\rm WS}$
    (\dag Huang et al.~\cite{XHuangCVPR2017},
    \ddag Yang et al.~\cite{JYangICLR2017})}
  \label{tab:quantitative}
\end{table}

\subsection{Combination with WGAN-GP}
\label{subsec:wgan-gp}

Another concern is whether our contributions are orthogonal
to the state-of-the-art GAN training techniques.
To demonstrate this,
we tested the {\bf DTLC$^L$-WGAN-GP} on three cases:
CIFAR-10 (unsupervised/weakly supervised) and
Tiny ImageNet\footnote{Tiny version of the ImageNet dataset
  containing 200 classes $\times$ 500 images.
  To shorten the training time, we resized images to $32 \times 32$.}
(unsupervised).
The number of categories was same as
that with the models used in Table~\ref{tab:quantitative}.
We list the results in Table~\ref{tab:inception_score}.
Interestingly, in all cases, the scores improved
as the layers became deeper, and
the {\bf DTLC$^4$-WGAN-GPs} achieved state-of-the-art performance.
We show generated image samples in
Figures~\ref{fig:cifar10_all_wgangp_unsup_ex}--\ref{fig:imagenet_all_wgangp_unsup_ex} of the appendix.

\begin{table}[tb]
  \begin{center}
    \footnotesize{
      \begin{tabular*}{\columnwidth}{lccc}
        & {\bf CIFAR-10}
        & {\bf CIFAR-10}
        & {\bf Tiny ImageNet} \\
        {\bf Model}
        & \!\!\!{\bf (Unsupervised)}\!\!\!
        & \!\!\!{\bf (Supervised)}\!\!\!
        & \!\!\!{\bf (Unsupervised)}\!\!\!
        \\ \Xhline{0.8pt}        
        \!\!\!WGAN-GP
        & 7.86 $\pm$ .07\dag\!\!\!
        & -
        & 8.33 $\pm$ .11 \\
        \!\!\!AC/Info-WGAN-GP
        & 7.97 $\pm$ .09
        & 8.42 $\pm$ .10\dag\!\!\!
        & 8.33 $\pm$ .10 \\
        \!\!\!DTLC$^2$-WGAN-GP
        & 8.03 $\pm$ .12
        & 8.44 $\pm$ .10
        & 8.34 $\pm$ .08 \\
        \!\!\!DTLC$^3$-WGAN-GP
        & 8.15 $\pm$ .08
        & 8.56 $\pm$ .07
        & 8.41 $\pm$ .10 \\
        \!\!\!DTLC$^4$-WGAN-GP
        & {\bf 8.22} $\pm$ {\bf .11}
        & {\bf 8.80} $\pm$ {\bf .08}
        & {\bf 8.51} $\pm$ {\bf .08} \\
        \hline
        \!\!\!State-of-the-Art
        & 7.86 $\pm$ .07\dag\!\!\!
        & 8.59 $\pm$ .12\ddag\!\!\!
        & - \\
      \end{tabular*}
    }
  \end{center}
  \vspace{-2mm}
  \caption{Inception scores for WGAN-GP-based models
    (\dag Gulrajani et al.~\cite{IGulrajaniNIPS2017},
    \ddag Huang et al.~\cite{XHuangCVPR2017})}
  \label{tab:inception_score}
\end{table}

\subsection{Extension to Continuous Codes}
\label{subsec:continuous}
\begin{figure*}[t]
\begin{center}
  \includegraphics[width=1.0\textwidth]{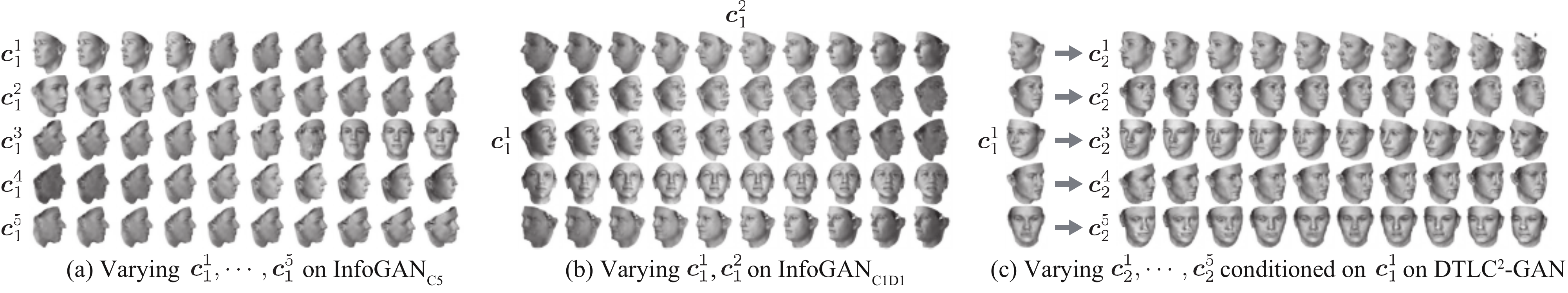}
\end{center}
\vspace{-2mm}
\caption{{\bf Representation comparison of models that have continuous codes:}
  Each sample is generated from same noise but different continuous codes
  (varied from left to right).
  In InfoGANs (a, b), each code is independent and exclusive, while
  in DTLC-GAN (c), lower layer codes learn category-specific
  (in this case, pose-specific) semantic features
  conditioned on higher layer codes.}
\label{fig:facegen}
\end{figure*}

To analyze the DTLC-GAN with continuous codes,
we evaluated it on the 3D Faces dataset,
which consists of faces generated from a 3D face model
and contains 240,000 samples.
We compared three models,
the {\bf InfoGAN$_{\rm C5}$},
which is the InfoGAN with five continuous codes
${\bm c}_1^1, \cdots, {\bm c}_1^5 \sim {\rm Unif} (-1, 1)$
(used in the InfoGAN study \cite{XChenNIPS2016}),
{\bf InfoGAN$_{\rm C1D1}$},
which is the InfoGAN with one categorical code
${\bm c}_1^1 \sim {\rm Cat}(K = 5, p=0.2)$ and
one continuous code
${\bm c}_1^2 \sim {\rm Unif}(-1, 1)$, and
{\bf DTLC$^2$-GAN}, which has one categorical code
${\bm c}_1^1 \sim {\rm Cat}(K = 5, p=0.2)$ in the first layer and
five continuous codes
${\bm c}_2^1, \cdots, {\bm c}_2^5 \sim {\rm Unif} (-1, 1)$
in the second layer.
We show example results in Figure~\ref{fig:facegen}.
In the {\bf InfoGANs} (a, b),
the individual codes tend to represent
independent and exclusive semantic features
because they have a flat relationship, while
in the {\bf DTLC$^2$-GAN} (c),
we can learn category-specific (in this case, pose-specific) semantic features
conditioned on the higher layer codes.

\subsection{Application to Image Retrieval}
\label{subsec:retrieval}
\begin{figure*}[t]
\begin{center}
  \includegraphics[width=1.0\textwidth]{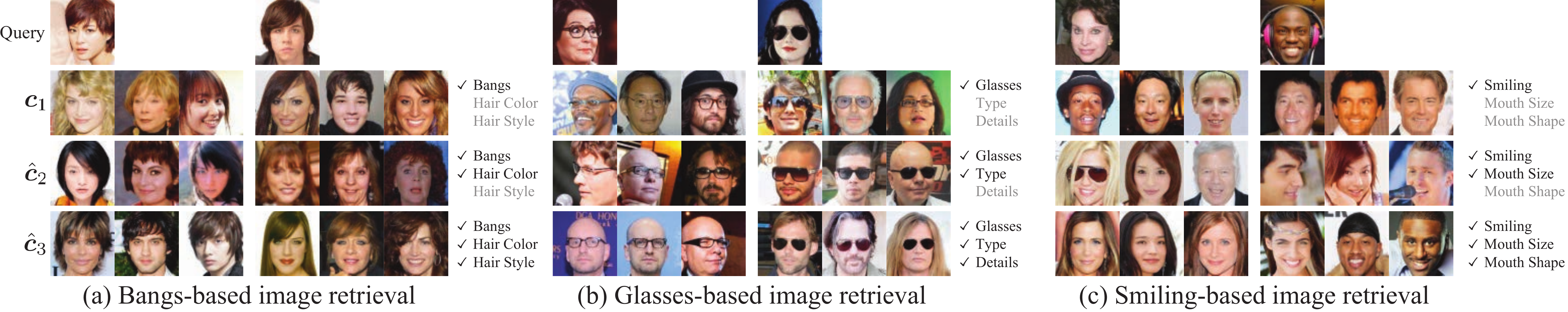}
\end{center}
\vspace{-2mm}
\caption{{\bf Example results of hierarchically interpretable image retrieval}:
  To search for image hierarchically, we measure L2 distance
  between query and database images on basis of ${\bm c}_1$, $\hat{\bm c}_2$,
  and $\hat{\bm c}_3$}
\label{fig:celeba_ret}
\end{figure*}

\begin{figure}[t]
\begin{center}
  \includegraphics[width=0.7\columnwidth]{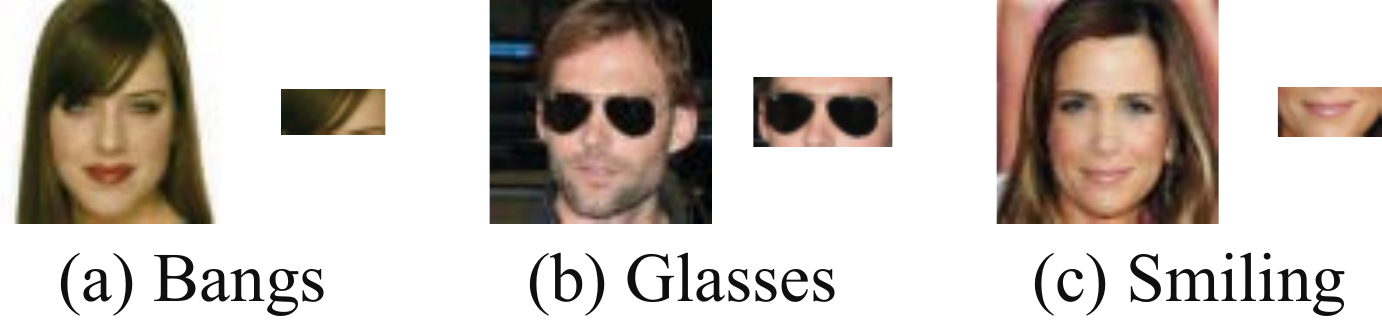}
\end{center}
\vspace{-2mm}
\caption{Attribute-specific areas used for evaluation in Table~\ref{tab:eval_ret}}
\label{fig:celeba_attr_area}
\end{figure}

One possible application of the DTLC-GAN is to use hierarchically
interpretable representations for image retrieval.
To confirm this, we used the CelebA dataset,
which consists of photographs of faces
and contains 180,000 training and 20,000 test samples.
To search for an image hierarchically,
we measure the L2 distance between query and database images
on the basis of ${\bm c}_1$, $\hat{\bm c}_2, \cdots, \hat{\bm c}_L$,
which are predicted using auxiliary functions $Q_1, \cdots, Q_L$.
Figure~\ref{fig:celeba_ret} shows the results of
bangs-based, glasses-based, and smiling-based image retrieval.
For evaluation, we used the test set in the CelebA dataset.
We trained {\bf DLTC$^3$-GAN$_{\rm WS}$},
where $k_1 = 2$, $k_2 = 3$, and $k_3 = 3$,
particularly where
hierarchical representations are learned only
for the attribute presence state.\footnote{We provide
  generated image samples in
  Figures~\ref{fig:celeba_bangs}--\ref{fig:celeba_smiling} of the appendix.}
These results indicate that
as the layer becomes deeper,
images in which attribute details match more can be retrieved.
To evaluate quantitatively, we measured the SSIM score
between query and database images
for the attribute-specific areas \cite{TKanekoCVPR2017}
defined in Figure~\ref{fig:celeba_attr_area}.
We summarize the scores in Table~\ref{tab:eval_ret}.
These results indicate that as the layer becomes deeper,
the concordance rate of attribute-specific areas increases.

\begin{table}
  \begin{center}
    \footnotesize{
      \begin{tabular}{lccc}
        {\bf Code} & {\bf Bangs} & {\bf Glasses} & {\bf Smiling}
        \\ \Xhline{0.8pt}
        ${\bm c}_1$ & 0.150 & 0.189 & 0.274 \\
        $\hat{\bm c}_2$ & 0.194 & 0.256 & 0.294 \\
        $\hat{\bm c}_3$ & {\bf 0.211} & {\bf 0.265} & {\bf 0.326} \\
      \end{tabular}
    }
  \end{center}
  \vspace{-2mm}
  \caption{Attribute-specific SSIM scores for different codes}
  \label{tab:eval_ret}
\end{table}

\section{Discussion and Conclusions}
\label{sec:discussion_conclusion}

We proposed an extension of the GAN called the DTLC-GAN
to learn hierarchically interpretable representations.
To develop it,
we introduced the DTLC to impose a hierarchical inclusion structure
on latent variables
and proposed the HCMI and curriculum learning method
to discover the salient semantic features in a layer-by-layer manner
by only using a single DTLC-GAN model
without relying on detailed supervision.
Experiments showed promising results, indicating that
the DTLC-GAN
is well suited for learning hierarchically interpretable representations.
The DTLC-GAN is a general model, and
possible future work includes
applying it to other models, such as encoder-decoder models
\cite{JDonahueICLR2017,VDumoulinICLR2017,DKingmaICLR2014,ALarsenICML2016,DRezendeICML2014},
and using it as a latent hierarchical structure discovery tool
for high-dimensional data.

\clearpage

{\small
\bibliographystyle{ieee}
\bibliography{refs}
}

\clearpage

\appendix

In this appendix,
we provide
additional analysis in Section~\ref{sec:analysis},
give details on the experimental setup in Section~\ref{sec:detail},
and provide extended results in Section~\ref{sec:extend}.
We provide other supplementary materials including demo videos
at \url{http://www.kecl.ntt.co.jp/people/kaneko.takuhiro/projects/dtlc-gan/}.

\section{Additional Analysis}
\label{sec:analysis}

\subsection{Representation Comparison on Simulated Data}
\label{subsec:simulation}
To clarify the limitation of the InfoGANs compared in
Figure~\ref{fig:representation_comparison},
we conducted experiments on simulated data.
In particular, we used simulated data
that are hierarchically sampled in the 2D space
and have globally ten categories and locally two categories.
When sampling data,
we first randomly selected a global position from ten candidates
that are equally spaced around a circle of radius $2$.
We then randomly selected
a local position from two candidates
that are rotated by $0.05$ radians in clockwise and anticlockwise directions
from the global position.
Based on this local position, we sampled data from a Gaussian distribution
of a standard deviation of $0.1$.

We compared models that are similar to those
compared in Figure~\ref{fig:representation_comparison}.
As the proposed model,
we used the {\bf DTLC$^2$-GAN}, where $k_1 = 10$ and $k_2 = 2$.
In this model, $\hat{\bm c}_2$, the dimension of which is
$10 \times 2 = 20$, is given to the $G$.
For comparison, we used two models in which latent code dimensions are also 20
but not hierarchical.
One is the {\bf InfoGAN$_{1 \times 20}$}, which has
one code ${\bm c}_1^1 \sim {\rm Cat}(K = 20, p = 0.05)$, and
the other is the {\bf InfoGAN$_{2 \times 10}$}, which has two codes
${\bm c}_1^1, {\bm c}_1^2 \sim {\rm Cat}(K = 10, p = 0.1)$.
For {\bf DTLC$^2$-GAN},
we also compared the {\bf DTLC$^2$-GANs} with and without curriculum learning.

\begin{figure*}[h]
  \begin{center}
    \includegraphics[width=0.9\textwidth]{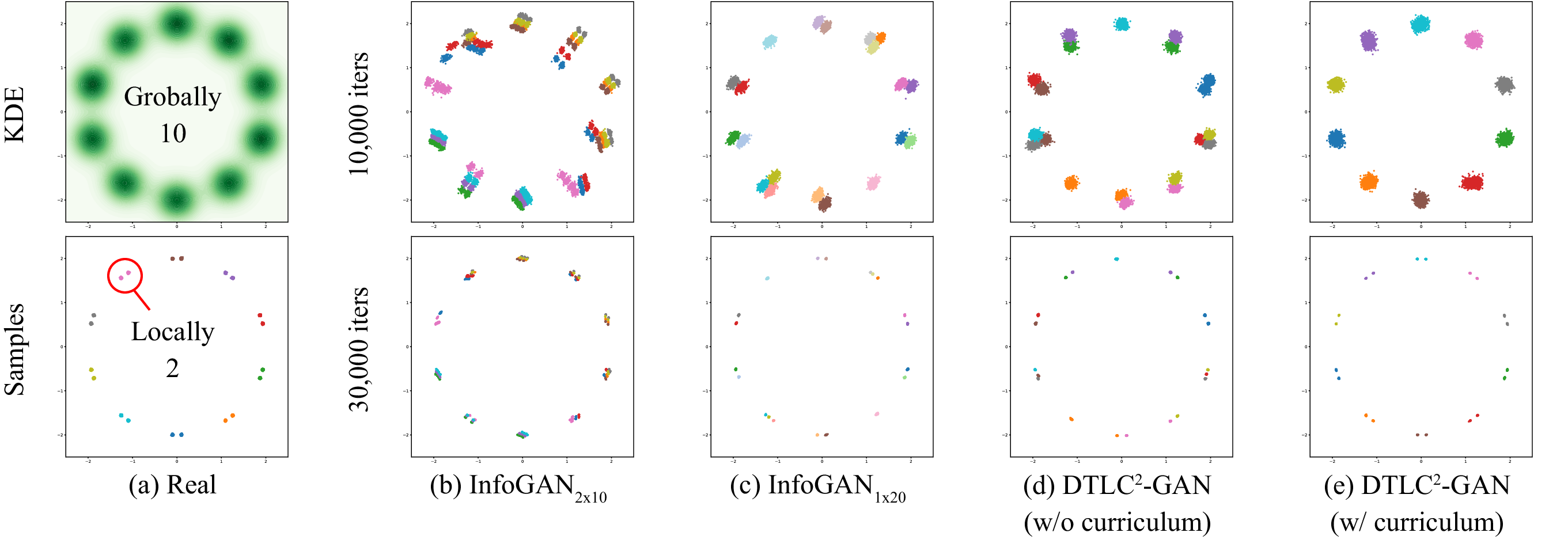}
  \end{center}
  \caption{{\bf Evaluation on simulated data:}
    (a) We used simulated data,
    which have globally ten categories and locally two categories.
    In (b), $10 \times 10 = 100$ categories are learned at the same time.
    In (c)(d), $20$ categories are learned at the same time,
    causing unbalanced and non-hierarchical clustering.
    In (e), ten global categories are first discovered
    then two local categories are learned.
    Upper left: kernel density estimation (KDE) plots.
    Others: samples from real data or models. Same color indicates same
    ${\bm c}_1^1$ category.}
  \label{fig:simulation}
\end{figure*}

We show the results in Figure~\ref{fig:simulation}.
The results indicate that
{\bf InfoGANs} (b, c) and {\bf DTLC$^2$-GAN} without curriculum learning (d)
tend to cause unbalanced or non-hierarchical clustering.
In contrast, the {\bf DTLC$^2$-GAN} with curriculum learning (e) succeeds in
capturing hierarchical structures, i.e.,
the first-layer codes captured global ten points, whereas
the second-layer codes captured local two points for each global position.

\subsection{Visual Interpretability Analysis}
\label{subsec:interpretability}
To clarity the benefit of learned representations,
we conducted two XAB tests.
For each test,
we compared the fourth-layer models
(DTLC$^4$-GANs or DTLC$^4$-WGAN-GPs)
with and without curriculum learning.

\begin{itemize}
\item {\bf Test I: Difference Interpretability Analysis}\\
To confirm whether $\hat{\bm c}_L$ is more interpretable than ${\bm z}$,
we compared the generated images (X) with
the images generated from latent variables
in which one dimension of ${\bm z}$ is changed (A)
and one dimension of $\hat{\bm c}_4$ is changed (B).
The changed dimension of ${\bm z}$ or $\hat{\bm c}_4$ was randomly chosen.
We asked participants which difference is more interpretable or even.

\item{\bf Test II: Semantic Similarity Analysis}\\
To confirm whether $\hat{\bm c}_L$ is hierarchically interpretable,
we compared the generated images (X) with
the images generated from latent variables
in which one dimension of ${\bm c}_2$ is varied (A)
and one dimension of ${\bm c}_4$ is varied (B).
For each case, we fixed the higher layer codes.
The changed dimension of ${\bm c}_2$ or ${\bm c}_4$ was randomly chosen.
The lower layer codes were also randomly chosen.
We asked participants which is semantically similar or even.
\end{itemize}

\begin{figure*}[thb]
  \begin{center}
    \includegraphics[width=\textwidth]{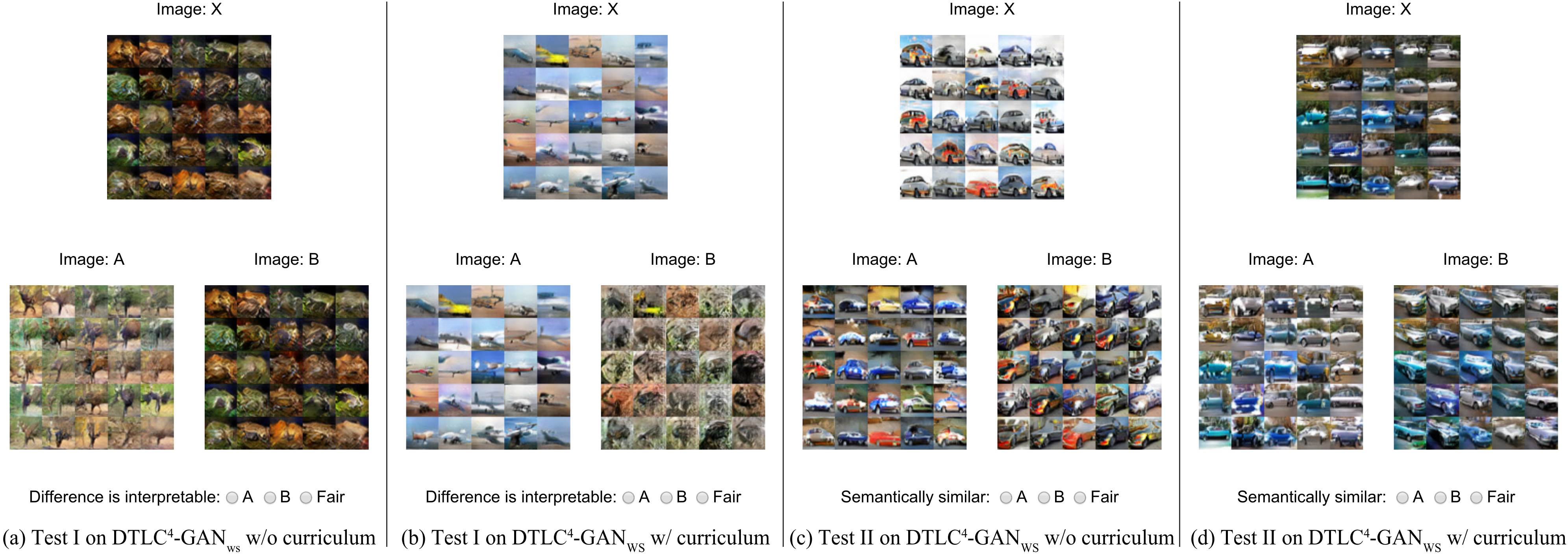}
  \end{center}
  \caption{{\bf User interfaces for XAB tests:}
    (a) Samples in ``Image: A'' are generated from latent variables
    in which one dimension of $\hat{\bm c}_L$ is changed.
    Samples in ``Image: B'' are generated from latent variables
    in which one dimension of ${\bm z}$ is changed.
    (b) Samples in ``Image: A'' are generated from latent variables
    in which one dimension of ${\bm z}$ is changed.
    Samples in ``Image: B'' are generated from latent variables
    in which one dimension of $\hat{\bm c}_L$ is changed.
    (c) Samples in ``Image: A'' are generated from latent variables
    in which ${\bm c}_4$ is varied.
    Samples in ``Image: B'' are generated from latent variables
    in which ${\bm c}_2$ is varied.
    (d) Samples in ``Image: A'' are generated from latent variables
    in which ${\bm c}_4$ is varied.
    Samples in ``Image: B'' are generated from latent variables
    in which ${\bm c}_2$ is varied.
  }
  \label{fig:xab_screenshot}
\end{figure*}

To eliminate bias in individual samples,
we showed 25 samples at the same time.
To eliminate bias in the order of stimuli,
the order (AB or BA) was randomly selected.
We show the user interfaces in Figure~\ref{fig:xab_screenshot}.

We summarize the results in Tables~\ref{tab:interpretability}.
In (a) and (b),
we list the results of tests I and II, respectively,
using the {\bf DTLC$^4$-GAN$_{\rm WS}$},
which were used for the experiments discussed in Figure~\ref{fig:cifar10_cd}.
The results of test I indicate that
$\hat{\bm c}_L$ is more interpretable than ${\bm z}$
regardless of curriculum learning.
We argue that this is because ${\bm z}$ does not have any constraints
on a structure and may be used by the $G$ in a highly entangled manner.
The results of test II indicate that
representations learned with curriculum learning
are hierarchically categorized in a better way
in terms of semantics than those without it.
The results support the effectiveness of the
proposed curriculum learning method.

We also conducted test II (semantic similarity analysis)
for all the {\bf DTLC$^4$-WGAN-GPs} discussed in Section~\ref{subsec:wgan-gp}.
We summarize the results in Table~\ref{tab:interpretability}(c)--(e).
We observed a similar tendency as those of {\bf DTLC$^4$-GAN$_{\rm WS}$}.

\subsection{Unsupervised Learning on Complex Dataset}
\label{subsec:unsupervised_ex}
\begin{figure*}[thb]
  \begin{center}
    \includegraphics[width=0.7\textwidth]{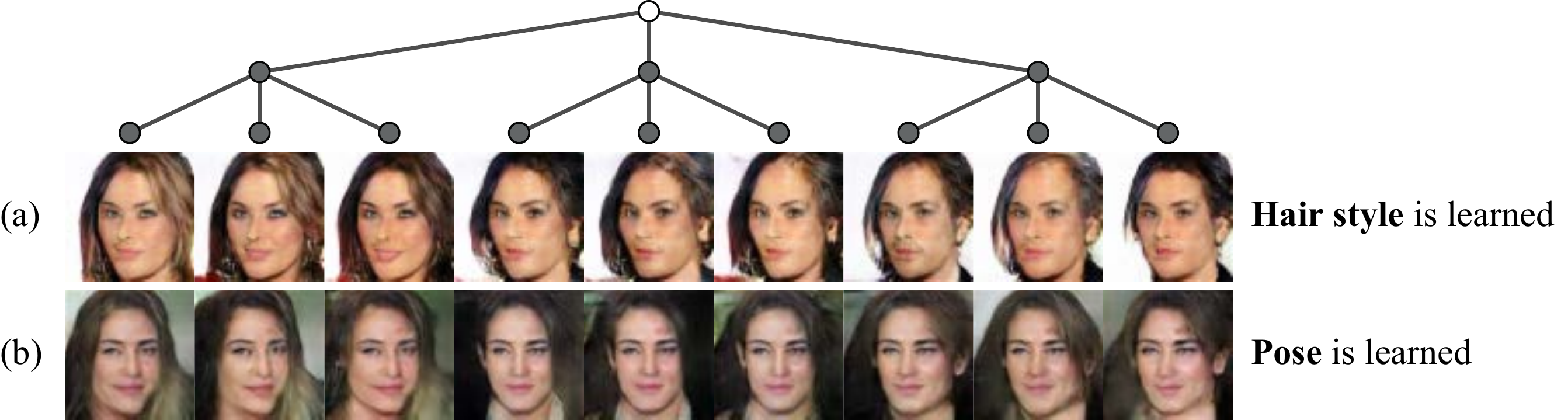}
  \end{center}
  \caption{{\bf Representation comparison between two models
    that are learned in fully unsupervised manner
    with different initialization:}
    In (a), samples are generated from one model, while,
    in (b), samples are generated from another model.
    In each row, ${\bm c}^1$ and ${\bm c}^2$ are varied
    per three images and per image, respectively.
    In this setting,
    learning targets (in (a), hair style and in (b), pose)
    depend on initialization because this dataset
    can be categorized in various ways.}
  \label{fig:celeba_unsup}
\end{figure*}

Although, in Section~\ref{subsec:unsupervised},
we mainly analyzed unsupervised settings on the MNIST dataset,
which is relatively simple,
we can learn hierarchical representations in an unsupervised manner
even in more complex datasets.
However, in this case, learning targets depend on the initialization
because such datasets can be categorized in various ways.
We illustrate this in Figure~\ref{fig:celeba_unsup}.
We also evaluated the DTLC-WGAN-GP in unsupervised settings
on the CIFAR-10 and Tiny ImageNet datasets.
See Section~\ref{subsec:wgan-gp} for details.

\begin{table}[t]
  \begin{center}
    \begin{small}
      \begin{tabular*}{\columnwidth}{lP{5em}P{5em}P{5em}}
        {\bf Model}
        & ${\bm z}$
        & even
        & $\hat{\bm c}_4$ \\
        \Xhline{0.8pt}
        W/o curriculum
        &  0.0
        &  1.0 $\pm$ 1.0
        & {\bf 99.0 $\pm$ 1.0} \\
        W/ curriculum
        &  0.0 
        &  1.0 $\pm$ 1.0
        & {\bf 99.0 $\pm$ 1.0} \\
        \multicolumn{4}{c}{\footnotesize{*Number of collected answers is 400}}
        \\
        \multicolumn{4}{c}{(a) Test I for DTLC$^4$-GAN$_{\rm WS}$ on CIFAR-10}
        \\ \\
        {\bf Model}
        & ${\bm c}_2$
        & even
        & ${\bm c}_4$ \\
        \Xhline{0.8pt}
        W/o curriculum
        & 22.4 $\pm$ 3.9
        & 41.3 $\pm$ 4.6
        & 36.2 $\pm$ 4.5\\
        W/ curriculum
        & 3.6 $\pm$ 1.7
        & 17.8 $\pm$ 3.5
        & {\bf 78.7 $\pm$ 3.8} \\
        \multicolumn{4}{c}{\footnotesize{*Number of collected answers is 450}}
        \\
        \multicolumn{4}{c}{(b) Test II for DTLC$^4$-GAN$_{\rm WS}$ on CIFAR-10}
        \\ \\        
        {\bf Model}
        & ${\bm c}_2$
        & even
        & ${\bm c}_4$ \\
        \Xhline{0.8pt}
        W/o curriculum
        & 18.0 $\pm$ 4.4
        & 31.3 $\pm$ 5.3
        & 50.7 $\pm$ 5.7 \\
        W/ curriculum
        & 4.7 $\pm$ 2.4
        & 12.0 $\pm$ 3.7
        & {\bf 83.3 $\pm$ 4.2} \\
        \multicolumn{4}{c}{\footnotesize{*Number of collected answers is 300}}
        \\
        \multicolumn{4}{c}{(c) Test II for DTLC$^4$-WGAN-GP on CIFAR-10}
        \\ \\
        {\bf Model}
        & ${\bm c}_2$
        & even
        & ${\bm c}_4$ \\
        \Xhline{0.8pt}
        W/o curriculum
        & 21.7 $\pm$ 4.7
        & 38.3 $\pm$ 5.5
        & 40.0 $\pm$ 5.6 \\
        W/ curriculum
        & 17.0 $\pm$ 4.3
        & 24.0 $\pm$ 4.9
        & {\bf 59.0 $\pm$ 5.6} \\
        \multicolumn{4}{c}{\footnotesize{*Number of collected answers is 300}}
        \\
        \multicolumn{4}{c}{(d) Test II for DTLC$^4$-WGAN-GP$_{\rm WS}$ on CIFAR-10}
        \\ \\
        {\bf Model}
        & ${\bm c}_2$
        & even
        & ${\bm c}_4$ \\
        \Xhline{0.8pt}
        W/o curriculum
        & 13.2 $\pm$ 4.2
        & 53.6 $\pm$ 6.2
        & 33.2 $\pm$ 5.9 \\
        W/ curriculum
        & 2.4 $\pm$ 1.9
        & 17.2 $\pm$ 4.7
        & {\bf 80.4 $\pm$ 5.0} \\
        \multicolumn{4}{c}{\footnotesize{*Number of collected answers is 250}}
        \\
        \multicolumn{4}{c}{(e) Test II on DTLC$^4$-WGAN-GP on Tiny ImageNet}
        \\ \\
      \end{tabular*}
    \end{small}
  \end{center}
  \vspace{-2mm}
  \caption{Average preference score (\%) with $95\%$ confidence intervals.
    We compared fourth-layer models (DTLC$^4$-GANs or DTLC$^4$-WGAN-GPs)
    with and without curriculum learning.}
  \label{tab:interpretability}
\end{table}

\clearpage

\clearpage

\begin{spacing}{0.99}
\section{Details on Experimental Setup}
\label{sec:detail}
In this section, we describe the network architectures and training scheme
for each dataset.
We designed the network architectures and training scheme
on the basis of techniques
introduced for the InfoGAN~\cite{XChenNIPS2016}.
The $D$ and $Q_1, \cdots, Q_L$ share all convolutional layers (Conv.), and
one fully connected layer (FC.) is added to the final layer
for $Q_1, \cdots, Q_L$.
This means that the difference in the calculation cost for the GAN and
DTLC-GAN is negligibly small.
For discrete code $\hat{\bm c}_l^m$,
we represented
$Q_l^m$ as softmax nonlinearity.
For continuous code $\hat{\bm c}_l^m$,
we parameterized $Q_l^m$ through a factored Gaussian.

In most of the experiments we conducted, we designed
the network architectures and training scheme on the basis of the techniques
introduced for the DCGAN~\cite{ARadfordICLR2016} and
did not use the state-of-the-art GAN training techniques
to evaluate whether the DTLC-GAN works well without relying on
such techniques.
To downscale and upscale,
we respectively used convolutions (Conv.~$\downarrow$) and
backward convolutions (Conv.~$\uparrow$),
i.e., fractionally strided convolutions,
with stride 2.
As activation functions,
we used rectified linear units
(ReLUs)~\cite{VNairICML2010} for the $G$,
while we used leaky rectified linear units
(LReLUs)~\cite{AMaasICML2013,BXuICMLW2015} for the $D$.
We applied batch normalization (BNorm)~\cite{SIoffeICML2015} to
all the layers except the generator output layer
and discriminator input layer.
We trained the networks using the Adam optimizer \cite{DPKingmaICLR2015}
with a minibatch of size $64$.
The learning rate was set to $0.0002$ for the $D/Q_1, \cdots, Q_L$
and to $0.001$ for the $G$.
The momentum term $\beta_1$ was set to $0.5$.

To demonstrate that
our contributions are orthogonal
to the state-of-the-art GAN training techniques,
we also tested the DTLC-WGAN-GP
(our DTLC-GAN with the WGAN-GP ResNet~\cite{IGulrajaniNIPS2017})
discussed in Section~\ref{subsec:wgan-gp}.
We used similar network architectures and training scheme
as the WGAN-GP ResNet,
except for the extended parts.

The details for each dataset are given below.

\subsection{MNIST}
\label{subsec:detail_mnist}

The DTLC$^L$-GAN network architectures for the MNIST dataset,
which were used for the experiments discussed in
Section~\ref{subsec:unsupervised},
are shown in
Table~\ref{tab:arch_mnist}.
As a pre-process,
we normalized the pixel value to the range $[0, 1]$.
In the generator output layers,
we used the Sigmoid function.
We used the {\bf DTLC$^L$-GAN}, where
$k_1 = 10$ and $k_2, \cdots, k_L =2$,
i.e., which has one discrete code ${\bm c}_1^1 \sim {\rm Cat}(K=10, p=0.1)$
in the first layer and
$N_l$ discrete codes ${\bm c}_l^n \sim {\rm Cat}(K=2, p=0.5)$
in the $l$th layer
where $l = (2, \cdots, L)$, $n = (1, \cdots, N_L)$,
and $N_l = \prod_{i=1}^{l-1} k_i$.
We added $\hat{\bm c}_L \in \mathbb{R}^{\prod_{l=1}^L k_l}$
to the generator input.
The trade-off parameters $\lambda_1, \cdots, \lambda_L$ were set to 0.1.
We trained the networks for $1 \times 10^4$ iterations in unsupervised settings.
As a curriculum for $\hat{\bm c}_l$ $(l = 2, \cdots, L)$,
we added regularization $- \lambda_l {\cal L}_{\rm HCMI}(G, Q_l)$ and sampling
after $2(l-1) \times 10^3$ iterations.

\subsection{CIFAR-10}
\label{subsec:detail_cifar10}

The DTLC$^L$-GAN network architectures for the CIFAR-10 dataset,
which were used for the experiments discussed in
Section~\ref{subsec:weaklysupervised},
are shown in
Table~\ref{tab:arch_cifar10}.
As a pre-process,
we normalized the pixel value to the range $[-1, 1]$.
In the generator output layers,
we used the Tanh function.
We used the {\bf DTLC$^L$-GAN$_{\rm WS}$}, where
$k_1 = 10$ and $k_2, \cdots, k_L =3$,
i.e., which has one ten-dimensional discrete code ${\bm c}_1^1$
in the first layer and
$N_l$ discrete codes ${\bm c}_l^n \sim {\rm Cat}(K=3, p=\frac{1}{3})$
in the $l$th layer
where $l = (2, \cdots, L)$, $n = (1, \cdots, N_l)$,
and $N_l = \prod_{i=1}^{l-1} k_i$.
We added $\hat{\bm c}_L \in \mathbb{R}^{\prod_{l=1}^L k_l}$
to the generator input.
We used the supervision (i.e., class labels) for ${\bm c}_1^1$.
The trade-off parameters $\lambda_1, \cdots, \lambda_L$ were set to 1.
We trained the networks for $1 \times 10^5$ iterations
in weakly supervised settings.
As a curriculum for $\hat{\bm c}_l$ $(l = 3, \cdots, L)$,
we added regularization $- \lambda_l {\cal L}_{\rm HCMI}(G, Q_l)$ and sampling
after $2(l-2) \times 10^4$ iterations.

\subsection{DTLC-WGAN-GP}
\label{subsec:detail_wgangp}

The DTLC$^L$-WGAN-GP network architectures
for the CIFAR-10 and Tiny ImageNet datasets,
which were used for the experiments discussed in Section~\ref{subsec:wgan-gp},
are similar to the WGAN-GP ResNet used
in a previous paper~\cite{IGulrajaniNIPS2017},
except for the extended parts.
We used the {\bf DTLC$^L$-WGAN-GP}, where
$k_1 = 10$ and $k_2, \cdots, k_L =3$,
i.e., which has one ten-dimensional discrete code ${\bm c}_1^1$
in the first layer and
$N_l$ discrete codes ${\bm c}_l^n \sim {\rm Cat}(K=3, p=\frac{1}{3})$
in the $l$th layer
where $l = (2, \cdots, L)$, $n = (1, \cdots, N_l)$,
and $N_l = \prod_{i=1}^{l-1} k_i$.
Following the AC-WGAN-GP ResNet implementation~\cite{IGulrajaniNIPS2017},
we used conditional batch normalization
(CBN)~\cite{HdVriesNIPS2017,VDumoulinICLR2017b}
to make the $G$ conditioned on the codes
$\hat{\bm c}_L \in \mathbb{R}^{\prod_{l=1}^L k_l}$.
CBN has two parameters,
i.e., gain parameter $\gamma_j$ and bias parameter $b_j$,
for each category, where $j = 1, \cdots, \prod_{l=1}^L k_l$.
As curriculum for sampling,
in learning the higher layer codes,
we used $\gamma_j$ and $b_j$ averaged
over those for the related lower layer node codes.

In unsupervised settings,
we sampled ${\bm c}_1^1$ from categorical distribution
${\bm c}_1^1 \sim {\rm Cat}(K=10, p=0.1)$.
The trade-off parameters $\lambda_1, \cdots, \lambda_L$ were set to 1.
We trained the networks for $1 \times 10^5$ iterations.
As a curriculum for $\hat{\bm c}_l$ $(l = 2, \cdots, L)$,
we added regularization $- \lambda_l {\cal L}_{\rm HCMI}(G, Q_l)$ and sampling
after $2(l-1) \times 10^4$ iterations.

In weakly supervised settings,
we used the supervision (i.e., class labels) for ${\bm c}_1^1$.
The $\lambda_1, \cdots, \lambda_L$ were set to 1.
We trained the networks for $1 \times 10^5$ iterations.
As a curriculum for $\hat{\bm c}_l$ $(l = 3, \cdots, L)$,
we added regularization $- \lambda_l {\cal L}_{\rm HCMI}(G, Q_l)$ and sampling
after $2(l-2) \times 10^4$ iterations.

\subsection{3D Faces}
\label{subsec:detail_3dfaces}

The DTLC$^L$-GAN network architectures for the 3D Faces dataset,
which were used for the experiments discussed in
Section~\ref{subsec:continuous},
are shown in
Table~\ref{tab:arch_3dfaces}.
As a pre-process,
we normalized the pixel value to the range $[0, 1]$.
In the generator output layers,
we used the Sigmoid function.
We used the {\bf DTLC$^2$-GAN}, where
$k_1 = 5$ and $k_2 = 1$,
i.e., which has one discrete code ${\bm c}_1^1 \sim {\rm Cat}(K=5, p=0.2)$
in the first layer and
five continuous codes ${\bm c}_2^1, \cdots, {\bm c}_2^5 \sim {\rm Unif}(-1, 1)$
in the second layer.
We added $\hat{\bm c}_2 \in \mathbb{R}^{\prod_{l=1}^2 k_l}$
to the generator input.
The trade-off parameters $\lambda_1$ and $\lambda_2$ were set to 1.
We trained the networks for $1 \times 10^4$ iterations in unsupervised settings.
As a curriculum for $\hat{\bm c}_2$,
we added regularization $- \lambda_2 {\cal L}_{\rm HCMI}(G, Q_2)$ and sampling
after $2 \times 10^3$ iterations.

\subsection{CelebA}
\label{subsec:detail_celeba}

The DTLC$^L$-GAN network architectures for the CelebA dataset,
which were used for the experiments discussed in Section~\ref{subsec:retrieval},
are shown in
Table~\ref{tab:arch_celeba}.
As a pre-process,
we normalized the pixel value to the range $[-1, 1]$.
In the generator output layers,
we used the Tanh function.
We used the {\bf DTLC$^L$-GAN$_{\rm WS}$}, where
$k_1 = 2$ and $k_2, \cdots, k_L =3$,
particularly where hierarchical representations are learned
only for the attribute presence state.
Therefore, $N_2 = 1$ and $N_l$ ($l = 3, \cdots, L$)
is calculated as $N_l = \prod_{i=2}^{l-1} k_i$.
This model has one two-dimensional discrete code
in the first layer and
$N_l$ discrete codes ${\bm c}_l^n \sim {\rm Cat}(K=3, p=\frac{1}{3})$
in the $l$th layer
where $l = (2, \cdots, L)$ and $n = (1, \cdots, N_l)$.
We added $\hat{\bm c}_L \in \mathbb{R}^{1 + \prod_{l=2}^L k_l}$
to the generator input.
We used the supervision (i.e., an attribute label) for ${\bm c}_1^1$.
The trade-off parameters $\lambda_1, \cdots, \lambda_L$ were set to
1, 0.1, and 0.04 for bangs, glasses, and smiling, respectively.
We trained the networks for $5 \times 10^4$ iterations
in weakly supervised settings.
As a curriculum for $\hat{\bm c}_l$ $(l = 3, \cdots, L)$,
we added regularization $- \lambda_l {\cal L}_{\rm HCMI}(G, Q_l)$ and sampling
after $2(l-2) \times 10^4$ iterations.

\subsection{Simulated Data}
\label{subsec:detail_simulation}

The DTLC$^L$-GAN network architectures for the simulated data
used for the experiments
discussed in Section~\ref{subsec:simulation},
are shown in Table~\ref{tab:arch_simulation}.
As a pre-process, we scaled the discriminator input
by factor 4 (roughly scaled to range $[-1, 1]$).
We used the {\bf DTLC$^2$-GAN}, where $k_1 = 10$ and $k_2 = 2$,
i.e., which has one discrete code ${\bm c}_1^1 \sim {\rm Cat}(K=10, p=0.1)$
in the first layer and
ten discrete codes
${\bm c}_2^1, \cdots, {\bm c}_2^{10} \sim {\rm Cat}(K=2, p=0.5)$
in the second layer.
We added $\hat{\bm c}_2 \in \mathbb{R}^{\prod_{l=1}^2 k_l}$
to the generator input.
The trade-off parameters $\lambda_1$ and $\lambda_2$ were set to 1.
We trained the networks using the Adam optimizer with a minibatch of size 512.
The learning rate was set to 0.0001 for $D/Q_1,Q_2$ and $G$.
The momentum term $\beta_1$ was set to 0.5.
We trained the networks for $3 \times 10^4$ iterations in unsupervised settings.
As a curriculum for $\hat{\bm c}_2$,
we added regularization $-\lambda_2 {\cal L}_{\rm HCMI}(G, Q_2)$ and sampling
after $2 \times 10^4$ iterations.
\end{spacing}

\newpage
\begin{table}[t]
  \begin{center}
    \begin{small}
      \begin{tabular}{l} \bhline{1pt}
        {\bf Generator} $G$ \\ \bhline{0.75pt}
        Input ${\bm z} \in \mathbb{R}^{64}$ $+$
        $\hat{\bm c}_L \in \mathbb{R}^{\prod_{l=1}^L k_l}$ \\ \hline
        1024 FC., BNorm, ReLU \\ \hline
        $7 \cdot 7 \cdot 128$ FC., BNorm, ReLU \\ \hline
        $4 \times 4$ 64 Conv. $\uparrow$, BNorm, ReLU \\ \hline
        $4 \times 4$ 1 Conv. $\uparrow$, Sigmoid \\ \bhline{1pt} \\ \bhline{1pt}

        {\bf Discriminator} $D$ /
        {\bf Auxiliary Function} $Q_1, \cdots, Q_L$ \\ \bhline{0.75pt}
        Input $28 \times 28$ 1 gray image \\ \hline
        $4 \times 4$ 64 Conv. $\downarrow$, LReLU \\ \hline
        $4 \times 4$ 128 Conv. $\downarrow$, BNorm, LReLU \\ \hline
        1024 FC., BNorm, LReLU \\ \hline
        FC. output for $D$ \\
        $[$128 FC., BNorm, LReLU$]$-FC. output for $Q_1, \cdots, Q_L$
        \\ \bhline{1pt}
      \end{tabular}
    \end{small}
  \end{center}
  \vspace{-2mm}
  \caption{DTLC$^L$-GAN network architectures used for MNIST}
  \label{tab:arch_mnist}
\end{table}

\begin{table}[h]
  \begin{center}  
    \begin{small}
      \begin{tabular}{l} \bhline{1pt}
        {\bf Generator} $G$ \\ \bhline{0.75pt}
        Input ${\bm z} \in \mathbb{R}^{128}$ $+$
        $\hat{\bm c}_L \in \mathbb{R}^{\prod_{l=1}^L k_l}$ \\ \hline
        $4 \cdot 4 \cdot 512$ FC., BNorm, ReLU \\ \hline
        $4 \times 4$ 256 Conv. $\uparrow$, BNorm, ReLU \\ \hline
        $4 \times 4$ 128 Conv. $\uparrow$, BNorm, ReLU \\ \hline
        $4 \times 4$ 64 Conv. $\uparrow$, BNorm, ReLU \\ \hline
        $3 \times 3$ 3 Conv., Tanh \\ \bhline{1pt} \\ \bhline{1pt}

        {\bf Discriminator} $D$ /
        {\bf Auxiliary Function} $Q_1, \cdots, Q_L$ \\ \bhline{0.75pt}
        Input $32 \times 32$ 3 color image \\ \hline
        $3 \times 3$ 64 Conv., LReLU, Dropout \\ \hline
        $4 \times 4$ 128 Conv. $\downarrow$, BNorm, LReLU, Dropout \\ \hline
        $3 \times 3$ 128 Conv., BNorm, LReLU, Dropout \\ \hline
        $4 \times 4$ 256 Conv. $\downarrow$, BNorm, LReLU, Dropout \\ \hline
        $3 \times 3$ 256 Conv., BNorm, LReLU, Dropout \\ \hline
        $4 \times 4$ 512 Conv. $\downarrow$, BNorm, LReLU, Dropout \\ \hline
        $3 \times 3$ 512 Conv., BNorm, LReLU, Dropout \\ \hline
        FC. output for $D$ \\
        $[$128 FC., BNorm, LReLU, Dropout$]$- \\
        \:\:\:\:\:\:\:\: FC. output for $Q_1, \cdots, Q_L$
        \\ \bhline{1pt}
      \end{tabular}
    \end{small}
  \end{center}
  \vspace{-2mm}
  \caption{DTLC$^L$-GAN network architectures used for CIFAR-10}
  \label{tab:arch_cifar10}
\end{table}

\newpage
\begin{table}[t]
  \begin{center}  
    \begin{small}
      \begin{tabular}{l} \bhline{1pt}
        {\bf Generator} $G$ \\ \bhline{0.75pt}
        Input ${\bm z} \in \mathbb{R}^{128}$ $+$
        $\hat{\bm c}_L \in \mathbb{R}^{\prod_{l=1}^L k_l}$ \\ \hline
        1024 FC., BNorm, ReLU \\ \hline
        $8 \cdot 8 \cdot 128$ FC., BNorm, ReLU \\ \hline
        $4 \times 4$ 64 Conv. $\uparrow$, BNorm, ReLU \\ \hline
        $4 \times 4$ 1 Conv. $\uparrow$, Sigmoid \\ \bhline{1pt} \\ \bhline{1pt}

        {\bf Discriminator} $D$ /
        {\bf Auxiliary Function} $Q_1, \cdots, Q_L$ \\ \bhline{0.75pt}
        Input $32 \times 32$ 1 gray image \\ \hline
        $4 \times 4$ 64 Conv. $\downarrow$, LReLU \\ \hline
        $4 \times 4$ 128 Conv. $\downarrow$, BNorm, LReLU \\ \hline
        1024 FC., BNorm, LReLU \\ \hline
        FC. output for $D$ \\
        $[$128 FC., BNorm, LReLU$]$-FC. output for $Q_1, \cdots, Q_L$
        \\ \bhline{1pt}
      \end{tabular}
    \end{small}
  \end{center}
  \vspace{-2mm}
  \caption{DTLC$^L$-GAN network architectures used for 3D Faces}
  \label{tab:arch_3dfaces}
\end{table}

\begin{table}[h]
  \begin{center}  
    \begin{small}
      \begin{tabular}{l} \bhline{1pt}
        {\bf Generator} $G$ \\ \bhline{0.75pt}
        Input ${\bm z} \in \mathbb{R}^{128}$ $+$
        $\hat{\bm c}_L \in \mathbb{R}^{1 + \prod_{l=2}^L k_l}$ \\ \hline
        $4 \cdot 4 \cdot 512$ FC., BNorm, ReLU \\ \hline
        $4 \times 4$ 256 Conv. $\uparrow$, BNorm, ReLU \\ \hline
        $4 \times 4$ 128 Conv. $\uparrow$, BNorm, ReLU \\ \hline
        $4 \times 4$ 64 Conv. $\uparrow$, BNorm, ReLU \\ \hline
        $4 \times 4$ 3 Conv. $\uparrow$, Tanh \\ \bhline{1pt} \\ \bhline{1pt}

        {\bf Discriminator} $D$ /
        {\bf Auxiliary Function} $Q_1, \cdots, Q_L$ \\ \bhline{0.75pt}
        Input $64 \times 64$ 3 color image \\ \hline
        $4 \times 4$ 64 Conv. $\downarrow$, LReLU \\ \hline
        $4 \times 4$ 128 Conv. $\downarrow$, BNorm, LReLU \\ \hline
        $4 \times 4$ 256 Conv. $\downarrow$, BNorm, LReLU \\ \hline
        $4 \times 4$ 512 Conv. $\downarrow$, BNorm, LReLU \\ \hline
        FC. output for $D$ \\
        $[$128 FC., BNorm, LReLU$]$-FC. output for $Q_1, \cdots, Q_L$
        \\ \bhline{1pt}
      \end{tabular}
    \end{small}
  \end{center}
  \vspace{-2mm}
  \caption{DTLC$^L$-GAN network architectures used for CelebA}
  \label{tab:arch_celeba}
\end{table}

\begin{table}[h]
  \begin{center}  
    \begin{small}
      \begin{tabular}{l} \bhline{1pt}
        {\bf Generator} $G$ \\ \bhline{0.75pt}
        Input ${\bm z} \in \mathbb{R}^{256}$ $+$
        $\hat{\bm c}_L \in \mathbb{R}^{\prod_{l=1}^L k_l}$ \\ \hline
        128 FC., ReLU \\ \hline
        128 FC., ReLU \\ \hline
        2 FC. \\ \bhline{1pt} \\ \bhline{1pt}

        {\bf Discriminator} $D$ /
        {\bf Auxiliary Function} $Q_1, \cdots, Q_L$ \\ \bhline{0.75pt}
        Input 2D simulated data \\
        (scaled by factor 4 (roughly scaled to range $[-1, 1]$)) \\ \hline
        128 FC. ReLU \\ \hline
        128 FC. ReLU \\ \hline
        FC. output for $D$ \\
        $[$128 FC., ReLU$]$-FC. output for $Q_1, \cdots, Q_L$
        \\ \bhline{1pt}
      \end{tabular}
    \end{small}
  \end{center}
  \vspace{-2mm}
  \caption{DTLC$^L$-GAN network architectures used for simulated data}
  \label{tab:arch_simulation}
\end{table}

\clearpage
\section{Extended Results}
\label{sec:extend}

\subsection{MNIST}
\label{subsec:extend_mnist}

\begin{figure*}[h]
\begin{center}
  \includegraphics[width=0.8\textwidth]{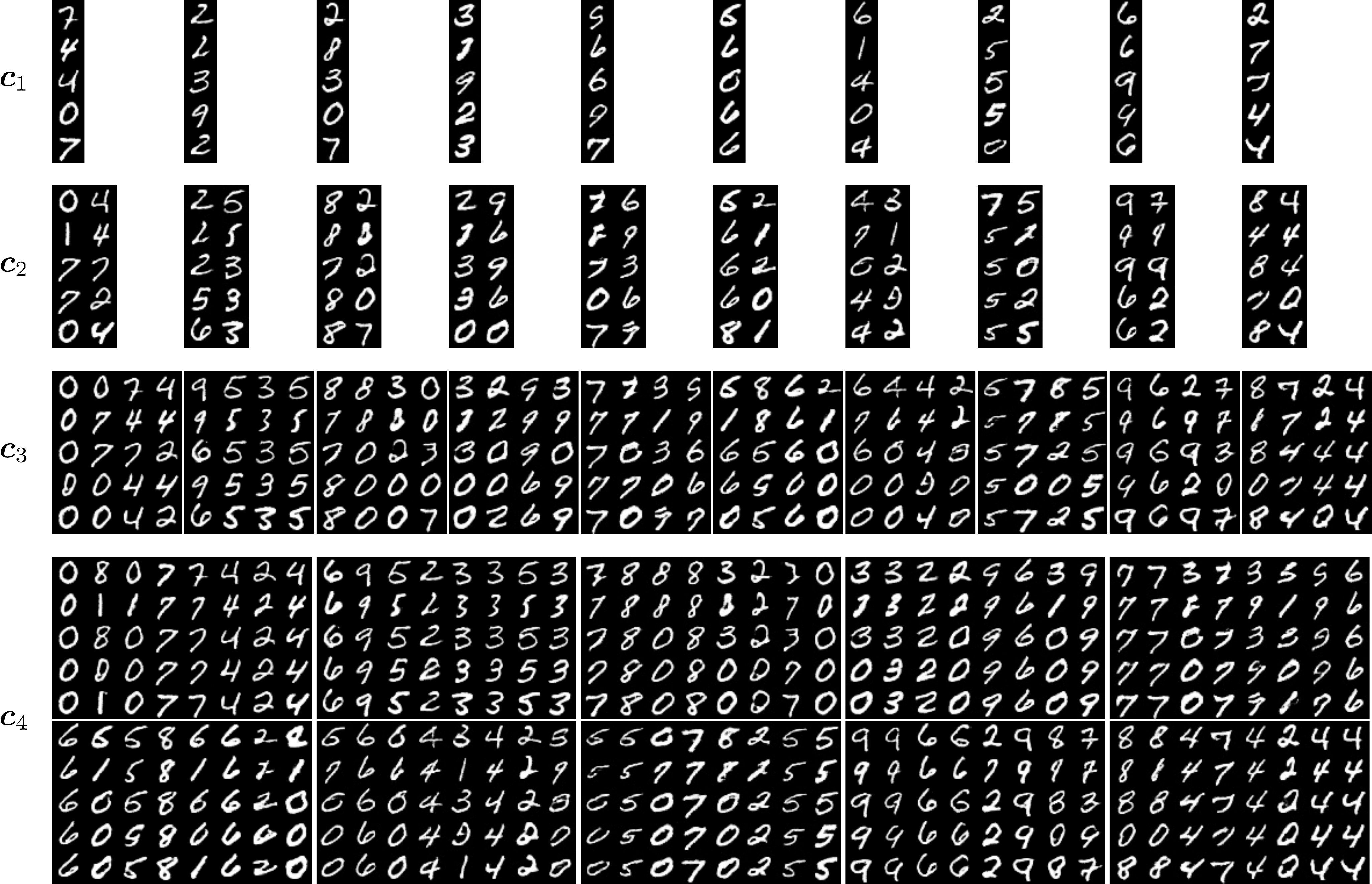}
\end{center}
\caption{{\bf Manipulating latent codes in
    DTLC$^4$-GAN (learned without curriculum) on MNIST:}
  In results noted with ${\bm c}_l$ $(l = 1, \cdots, 4)$,  
  each column includes five samples generated from  
  same ${\bm c}_{1}, \cdots, {\bm c}_l$
  but different ${\bm z}$
  and random ${\bm c}_{l+1}, \cdots, {\bm c}_L$.
  In each block, each row contains samples generated from
  same ${\bm z}$ and ${\bm c}_1$ but different
  ${\bm c}_2, \cdots, {\bm c}_L$.  
  In particular,
  ${\bm c}_i$ $(i = 2, \cdots, l-1)$
  was varied per $\prod_{j=i}^{l-1} k_j$ images, and
  ${\bm c}_l$ was varied per image.
  ${\bm c}_i$ $(i = l + 1, \cdots, L)$ was randomly chosen.}  
\label{fig:mnist_all_l0_s0_ex}
\end{figure*}

\begin{figure*}[h]
\begin{center}
  \includegraphics[width=0.8\textwidth]{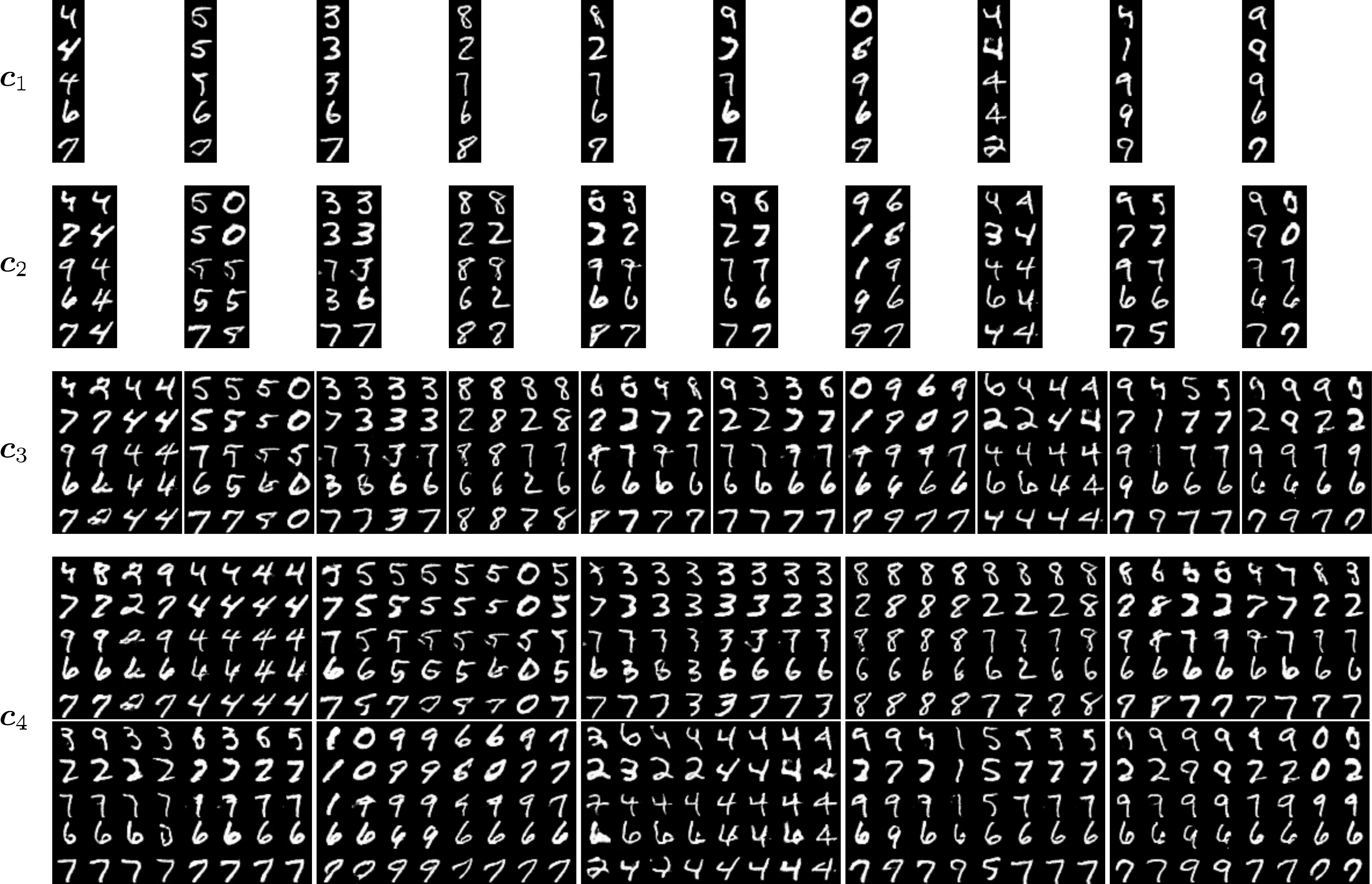}
\end{center}
\caption{{\bf Manipulating latent codes in
    DTLC$^4$-GAN (learned only with curriculum for regularization) on MNIST:}
  View of figure is same as that in Figure~\ref{fig:mnist_all_l0_s0_ex}}
\label{fig:mnist_all_l1_s0_ex}
\end{figure*}

\begin{figure*}[h]
\begin{center}
  \includegraphics[width=0.8\textwidth]{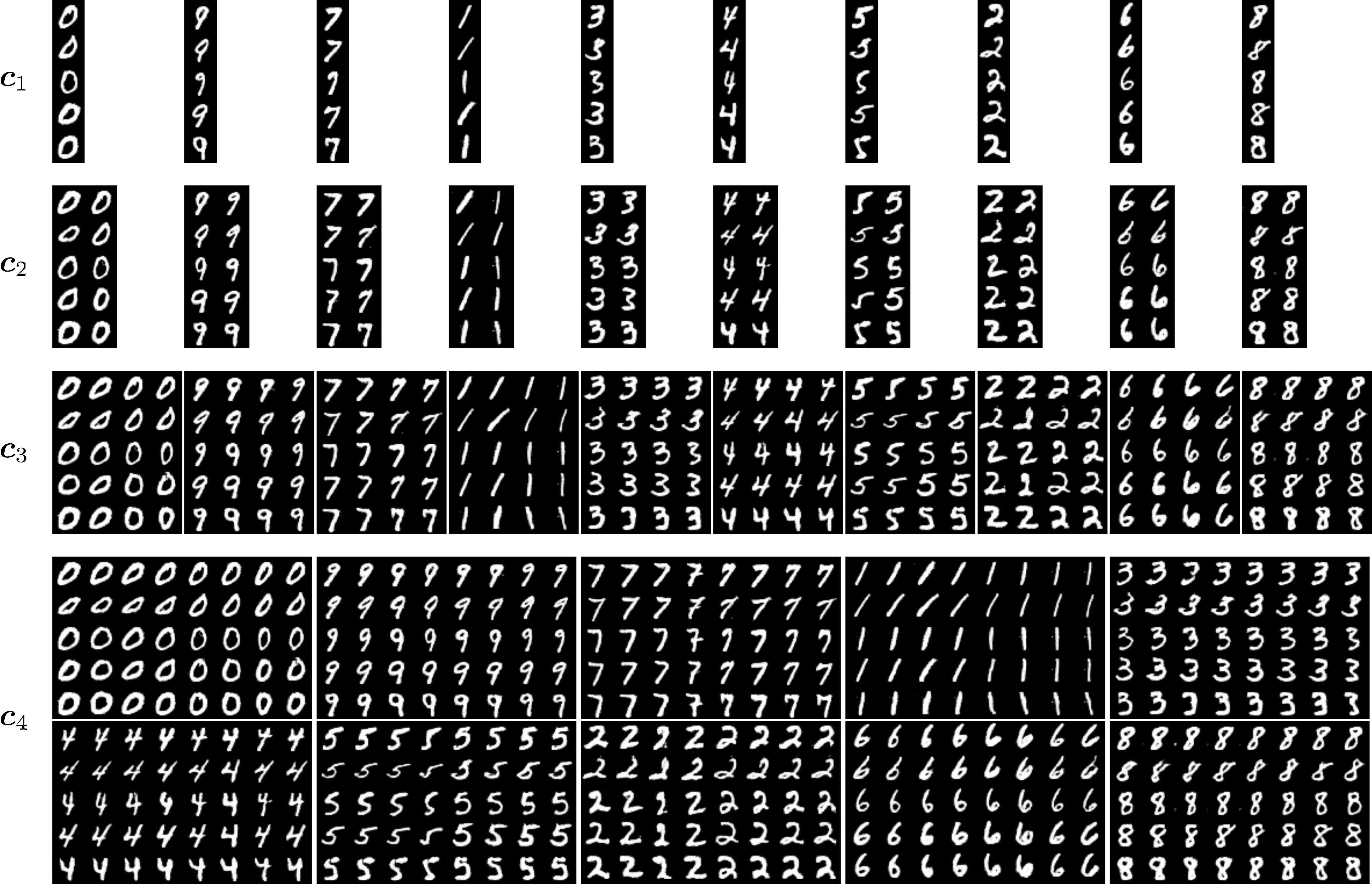}
\end{center}
\caption{{\bf Manipulating latent codes in
  DTLC$^4$-GAN (learned with full curriculum) on MNIST:}
  View of figure is same as that in Figure~\ref{fig:mnist_all_l0_s0_ex}}
\label{fig:mnist_all_ex}
\end{figure*}

We give extended results of Figure~\ref{fig:mnist_cd} in
Figures~\ref{fig:mnist_all_l0_s0_ex}--\ref{fig:mnist_all_ex}.
We used the {\bf DTLC$^4$-GAN},
where $k_1 = 10$ and $k_2, k_3, k_4 = 2$.
Figure~\ref{fig:mnist_all_l0_s0_ex} shows the generated image examples
using the {\bf DTLC$^4$-GAN} learned without a curriculum.
Figure~\ref{fig:mnist_all_l1_s0_ex} shows the generated image examples
using the {\bf DTLC$^4$-GAN}
learned only with the curriculum for regularization.
Figure~\ref{fig:mnist_all_ex} shows the generated image examples
using the {\bf DTLC$^4$-GAN} learned with the full curriculum
(curriculum for regularization and sampling:
proposed curriculum learning method).
The former two {\bf DTLC$^4$-GANs} (without the full curriculum) exhibited
confusion between inner-layer and intra-layer disentanglement, while
the {\bf DTLC$^4$-GAN} with the full curriculum succeeded in avoiding confusion.
The inner-category divergence evaluation on the basis of the SSIM
in Figure~\ref{fig:mnist_cd} also supports these observations.

\subsection{CIFAR-10}
\label{subsec:extend_cifar10}

\begin{figure*}[h]
\begin{center}
  \includegraphics[height=0.9\textheight]{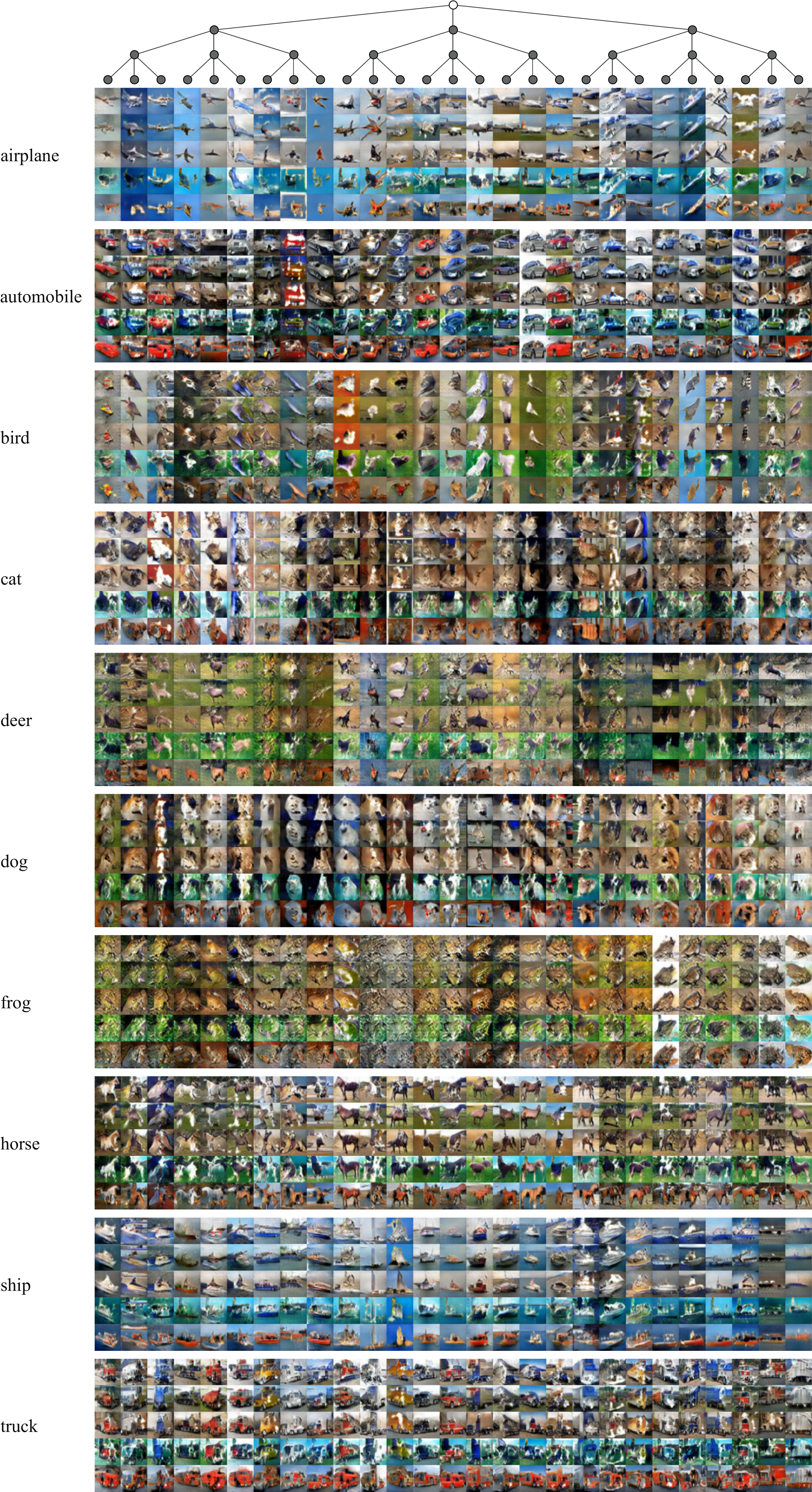}
\end{center}
\caption{{\bf Manipulating latent codes in
    DTLC$^4$-GAN$_{\rm WS}$ (learned without curriculum) on CIFAR-10:}  
  In each block,
  each column includes five samples generated from
  same ${\bm c}_1, \cdots, {\bm c}_4$
  but different ${\bm z}$.
  Each row contains samples generated from
  same ${\bm z}$ and ${\bm c}_1$ but
  different ${\bm c}_2$, ${\bm c}_3$, and ${\bm c}_4$.
  In particular,
  ${\bm c}_2$, ${\bm c}_3$, and ${\bm c}_4$ were varied
  per nine images, per three images, and per image, respectively.
  Among all blocks,
  samples in $i$th row $(i = 1, \cdots, 5)$ share same ${\bm z}$.}
\label{fig:cifar10_all_l0_s0_ex}
\end{figure*}

\begin{figure*}[h]
\begin{center}
  \includegraphics[height=0.9\textheight]{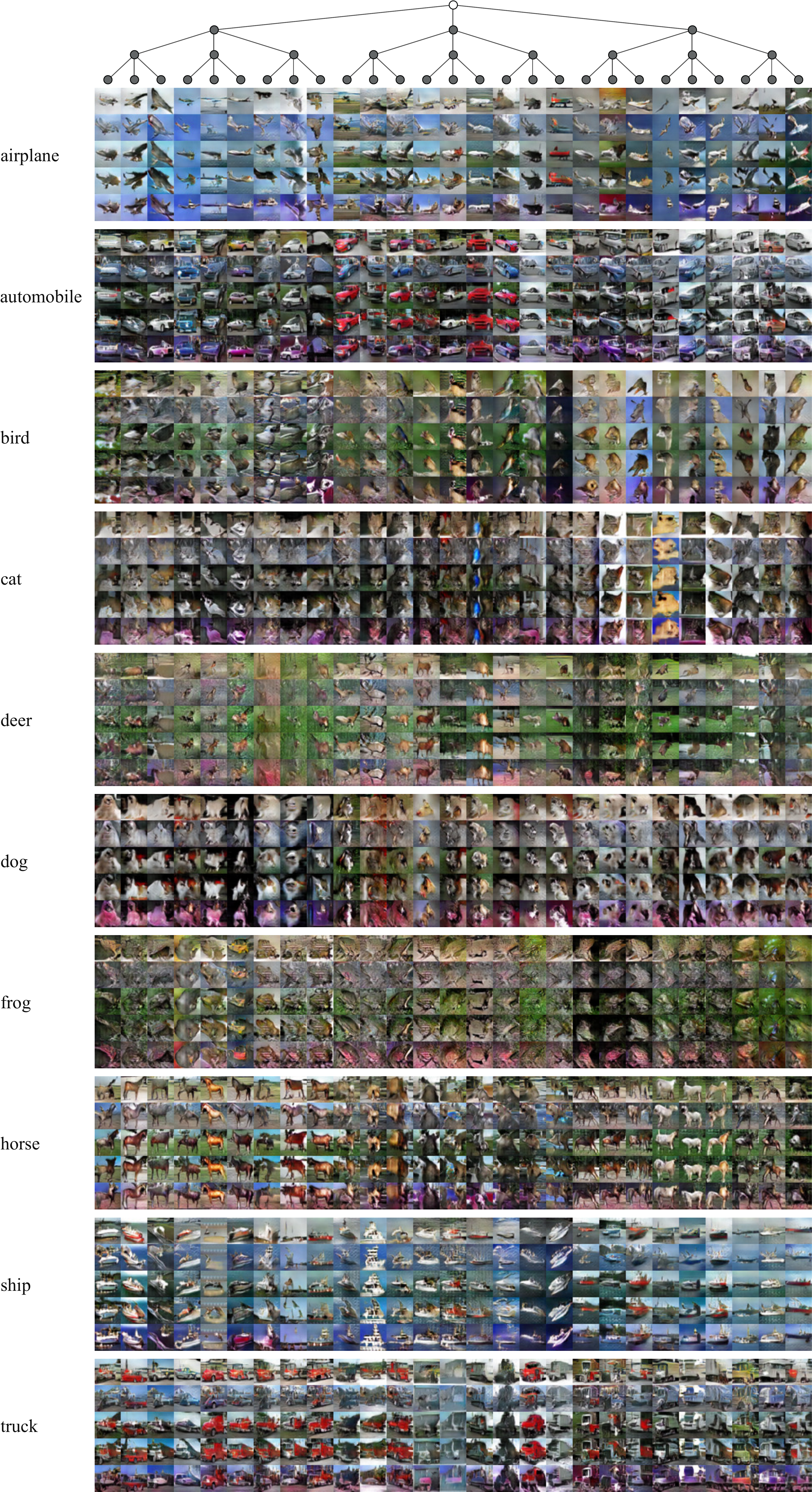}
\end{center}
\caption{{\bf Manipulating latent codes in
    DTLC$^4$-GAN$_{\rm WS}$ (learned only with curriculum for regularization)
    on CIFAR-10:}
  View of figure is same as that in Figure~\ref{fig:cifar10_all_l0_s0_ex}}
\label{fig:cifar10_all_l1_s0_ex}
\end{figure*}

\begin{figure*}[h]
\begin{center}
  \includegraphics[height=0.9\textheight]{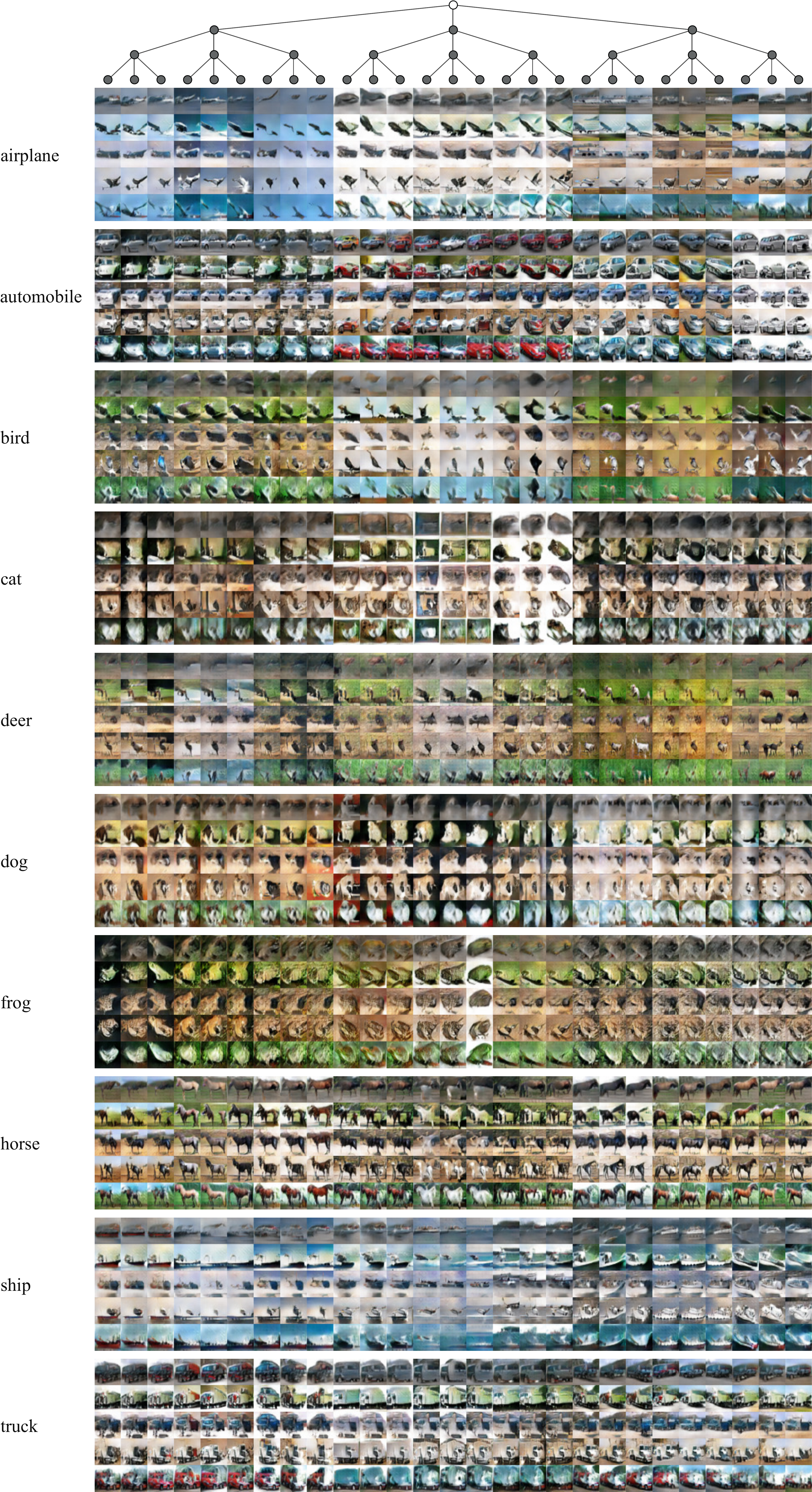}
\end{center}
\caption{{\bf Manipulating latent codes in
    DTLC$^4$-GAN$_{\rm WS}$ (learned with full curriculum) on CIFAR-10:}
  View of figure is same as that in Figure~\ref{fig:cifar10_all_l0_s0_ex}}
\label{fig:cifar10_all_ex}
\end{figure*}

We give extended results of Figure~\ref{fig:cifar10_cd} in
Figures~\ref{fig:cifar10_all_l0_s0_ex}--\ref{fig:cifar10_all_ex}.
We used the {\bf DTLC$^4$-GAN$_{\rm WS}$},
where $k_1 = 10$ and $k_2, k_3, k_4 = 3$.
We used class labels as supervision.
Figure~\ref{fig:cifar10_all_l0_s0_ex} shows the generated image samples
using the {\bf DTLC$^4$-GAN$_{\rm WS}$} learned without a curriculum.
Figure~\ref{fig:cifar10_all_l1_s0_ex} shows the generated image samples
using the {\bf DTLC$^4$-GAN$_{\rm WS}$} learned
only with the curriculum for regularization.
Figure~\ref{fig:cifar10_all_ex} shows the generated image samples
using the {\bf DTLC$^4$-GAN$_{\rm WS}$} learned with the full curriculum
(curriculum for regularization and sampling:
proposed curriculum learning method).
All models succeeded in learning disentangled representations in class labels
since they are given as supervision; however,
the former two {\bf DTLC$^4$-GAN$_{\rm WS}$s} (without the full curriculum)
exhibited confusion between inner-layer and intra-layer disentanglement
from second- to fourth-layer codes.
In contrast, the {\bf DTLC$^4$-GAN$_{\rm WS}$} with the full curriculum
succeeded in avoiding confusion.
The inner-category divergence evaluation on the basis of the SSIM
in Figure~\ref{fig:cifar10_cd} also supports these observations.

\subsection{DTLC-WGAN-GP}
\label{subsec:extend_wgangp}

\begin{figure*}[h]
\begin{center}
  \includegraphics[height=0.9\textheight]{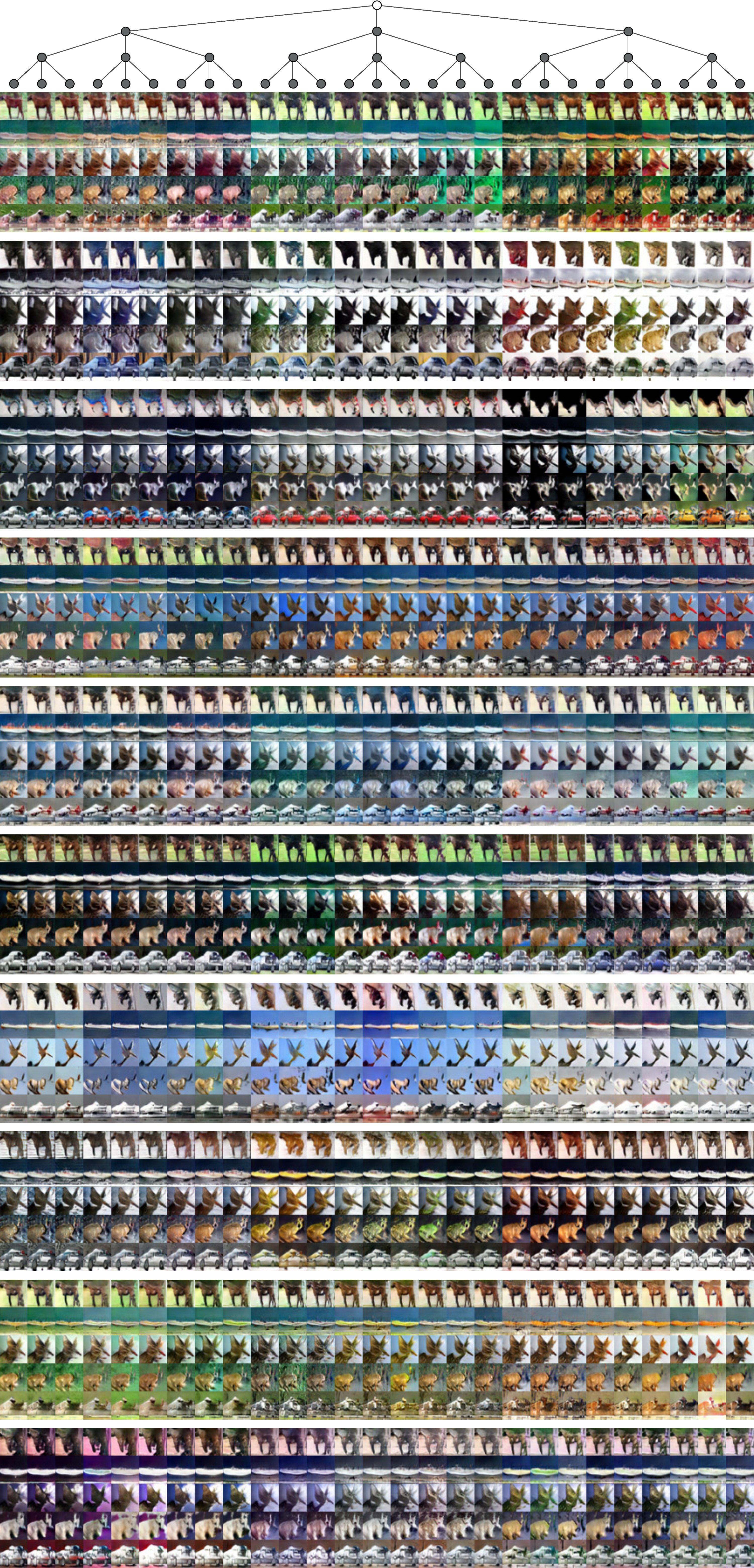}
\end{center}
\caption{{\bf Manipulating latent codes in
    DTLC$^4$-WGAN-GP on CIFAR-10:}
  View of figure is similar to that in Figure~\ref{fig:cifar10_all_l0_s0_ex}.
  All categories ($10 \times 3 \times 3 \times 3 = 270$) were learned
  in fully unsupervised setting.}
\label{fig:cifar10_all_wgangp_unsup_ex}
\end{figure*}

\begin{figure*}[h]
\begin{center}
  \includegraphics[height=0.9\textheight]{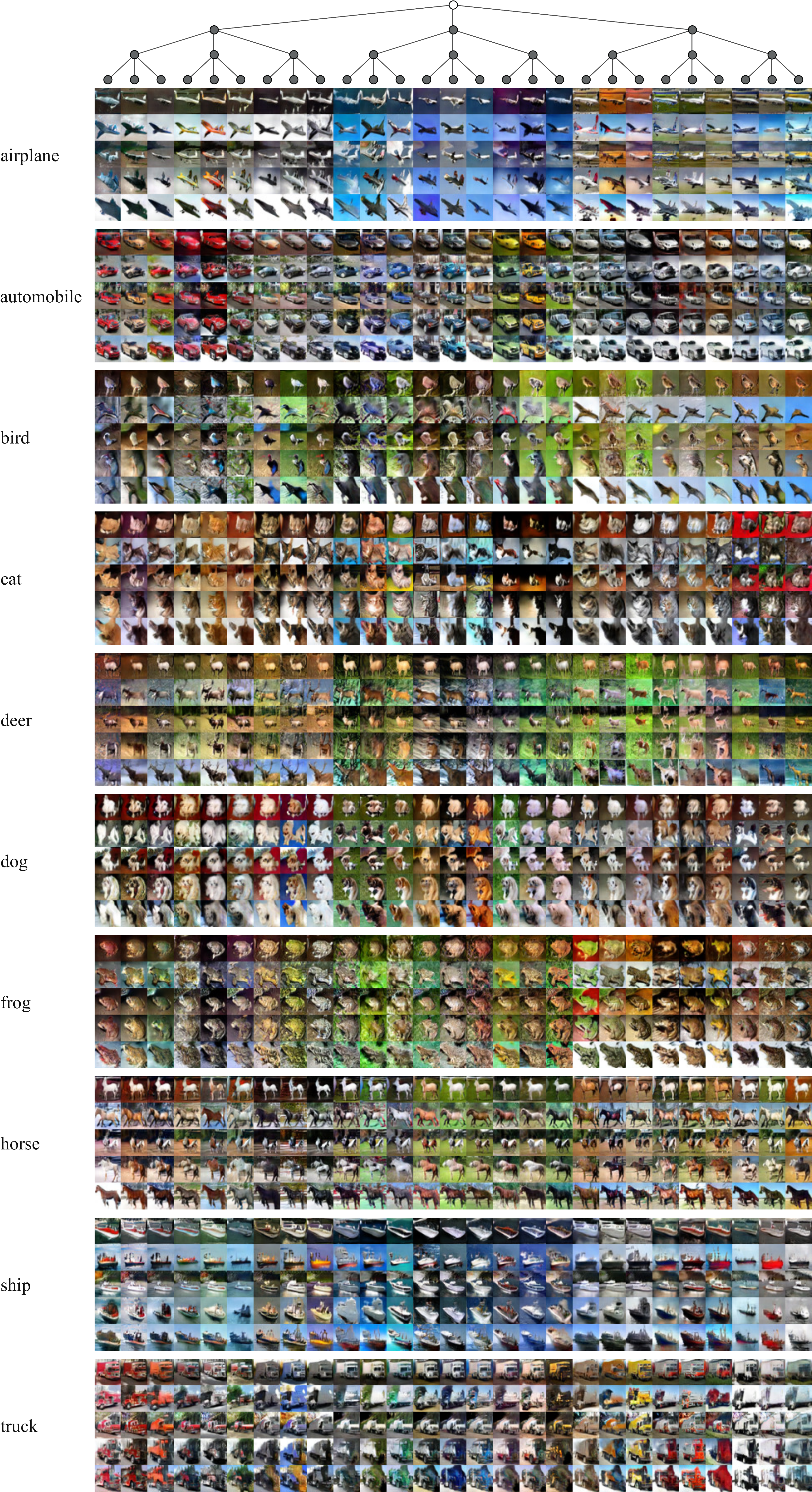}
\end{center}
\caption{{\bf Manipulating latent codes in
    DTLC$^4$-WGAN-GP$_{\rm WS}$ on CIFAR-10:}
  View of figure is similar to that in Figure~\ref{fig:cifar10_all_l0_s0_ex}.
  All categories ($10 \times 3 \times 3 \times 3 = 270$) were learned
  in weakly supervised setting.}
\label{fig:cifar10_all_wgangp_sup_ex}
\end{figure*}

\begin{figure*}[h]
\begin{center}
  \includegraphics[height=0.9\textheight]{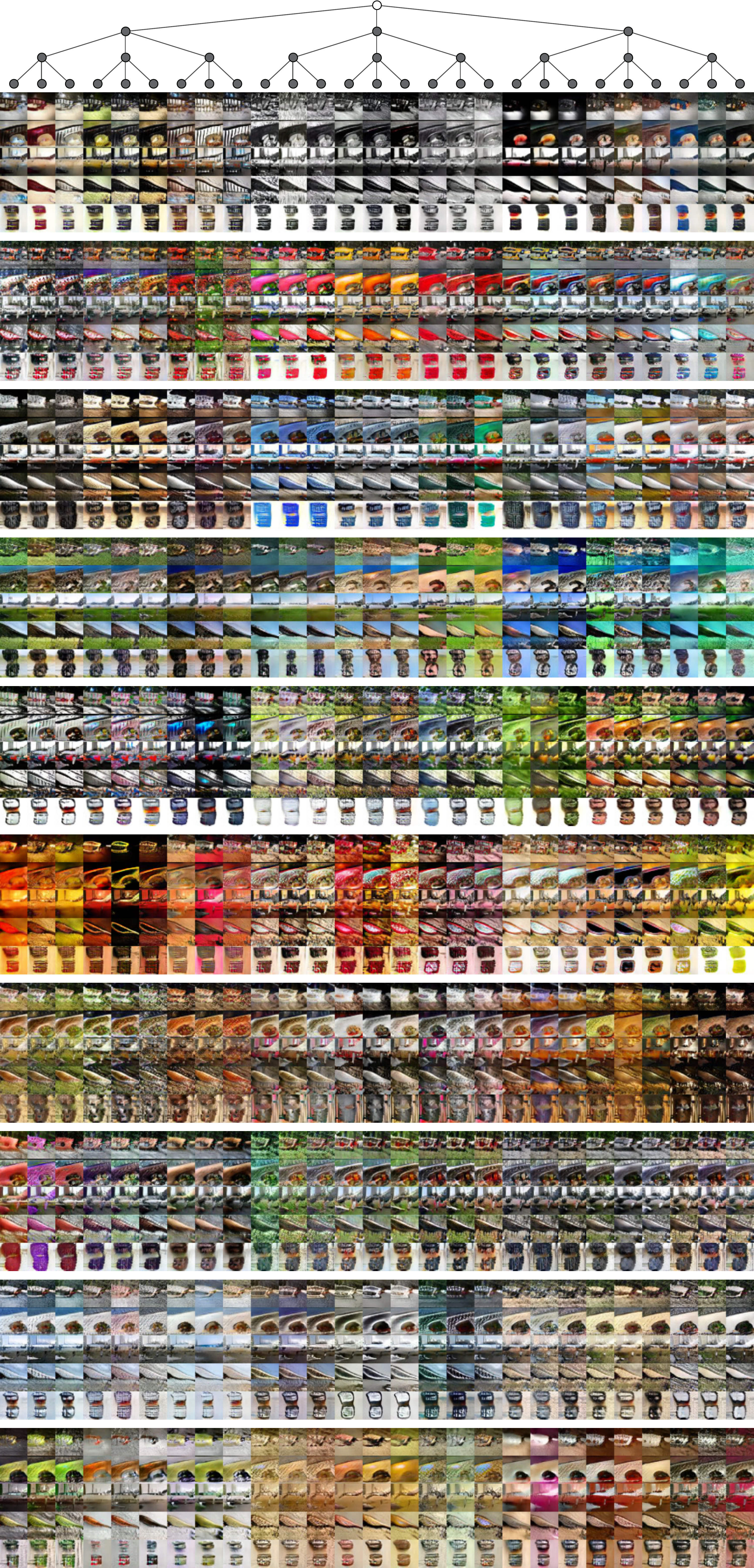}
\end{center}
\caption{{\bf Manipulating latent codes in
    DTLC$^4$-WGAN-GP on Tiny ImageNet:}
  View of figure is similar to that in Figure~\ref{fig:cifar10_all_l0_s0_ex}.
  All categories ($10 \times 3 \times 3 \times 3 = 270$) were learned
  in fully unsupervised setting.}
\label{fig:imagenet_all_wgangp_unsup_ex}
\end{figure*}

We show the generated image samples using the models
discussed in Section~\ref{subsec:wgan-gp},
in Figure~\ref{fig:cifar10_all_wgangp_unsup_ex}--\ref{fig:imagenet_all_wgangp_unsup_ex}.
We used the {\bf DTLC$^4$-WGAN-GP}, where $k_1 = 10$ and $k_2,k_3,k_4=3$.
In weakly supervised settings,
we used class labels as supervision.
Figure~\ref{fig:cifar10_all_wgangp_unsup_ex} shows the generated image samples
using the {\bf DTLC$^4$-WGAN-GP} on CIFAR-10 (unsupervised).
Figure~\ref{fig:cifar10_all_wgangp_sup_ex} shows the generated image samples
using the {\bf DTLC$^4$-WGAN-GP$_{\rm WS}$} on CIFAR-10 (weakly supervised).
Figure~\ref{fig:imagenet_all_wgangp_unsup_ex} shows the generated image samples
using the {\bf DTLC$^4$-WGAN-GP} on Tiny ImageNet (unsupervised).

\subsection{3D Faces}
\label{subsec:extend_3dfaces}

\begin{figure*}[h]
\begin{center}
  \includegraphics[width=\textwidth]{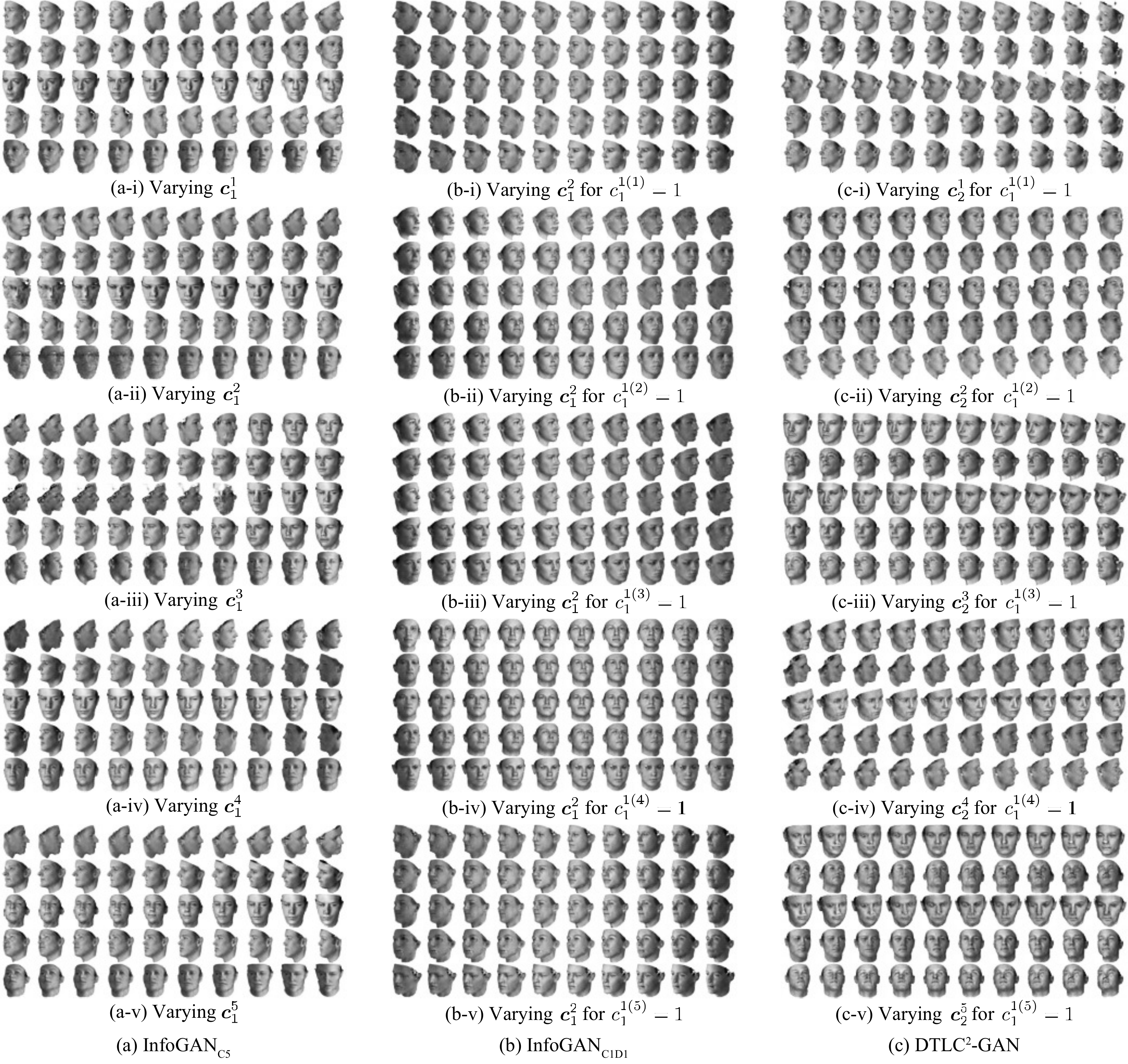}
\end{center}
\caption{{\bf Manipulating latent codes on 3D Faces:}
  In each block, each column includes five samples generated from
  same latent codes but different noise.
  Each row contains samples generated from same noise and same discrete codes
  but different continuous codes (varied from left tor right).
  In tandem blocks, samples in $i$th row ($i = 1, \cdots, 5$) were
  generated from same noise.
  (a) In {\bf InfoGAN$_{\rm C5}$},
  each continuous code ${\bm c}_1, \cdots, {\bm c}_5$ captures
  independent and exclusive semantic features
  (e.g., orientation of lighting in $c_1^4$ and elevation in $c_1^5$).
  (b) In {\bf InfoGAN$_{\rm C1D1}$},
  discrete code ${\bm c}_1^1$ captures pose, while
  continuous code ${\bm c}_1^2$ captures orientation of lighting
  regardless of ${\bm c}_1^1$.
  Also in this model,
  each code captures independent and exclusive semantic features.
  (c) In {\bf DTLC$^2$-GAN},
  discrete code ${\bm c}_1^1$ captures pose,  and
  continuous codes ${\bm c}_2^1, \cdots, {\bm c}_2^5$
  capture detailed variations for each pose.
  In this model, lower layer codes learn category-specific
  (in this case, pose-specific) semantic features conditioned on
  higher layer codes.}
\label{fig:facegen_all}
\end{figure*}

We give extended results of Figure~\ref{fig:facegen}
in Figure~\ref{fig:facegen_all}.
Similarly to Figure~\ref{fig:facegen},
we compared three models,
the {\bf InfoGAN$_{\rm C5}$},
which is the InfoGAN with five continuous codes
${\bm c}_1^1, \cdots, {\bm c}_1^5 \sim {\rm Unif} (-1, 1)$
(used in the InfoGAN study \cite{XChenNIPS2016}),
{\bf InfoGAN$_{\rm C1D1}$},
which is the InfoGAN with one categorical code
${\bm c}_1^1 \sim {\rm Cat}(K = 5, p=0.2)$ and
one continuous code
${\bm c}_1^2 \sim {\rm Unif}(-1, 1)$, and
{\bf DTLC$^2$-GAN}, which has one categorical code
${\bm c}_1^1 \sim {\rm Cat}(K = 5, p=0.2)$ in the first layer and
five continuous codes
${\bm c}_2^1, \cdots, {\bm c}_2^5 \sim {\rm Unif} (-1, 1)$
in the second layer.
In the {\bf InfoGAN$_{\rm C5}$} and {\bf InfoGAN$_{\rm C1D1}$},
the individual codes tend to capture
independent and exclusive semantic features
because they have a flat relationship, while
in the {\bf DTLC$^2$-GAN},
lower layer codes learn category-specific
(in this case, pose-specific) semantic features conditioned on
higher layer codes.

\subsection{CelebA}
\label{subsec:extend_celeba}
\begin{figure*}
\begin{center}
  \includegraphics[width=0.55\textwidth]{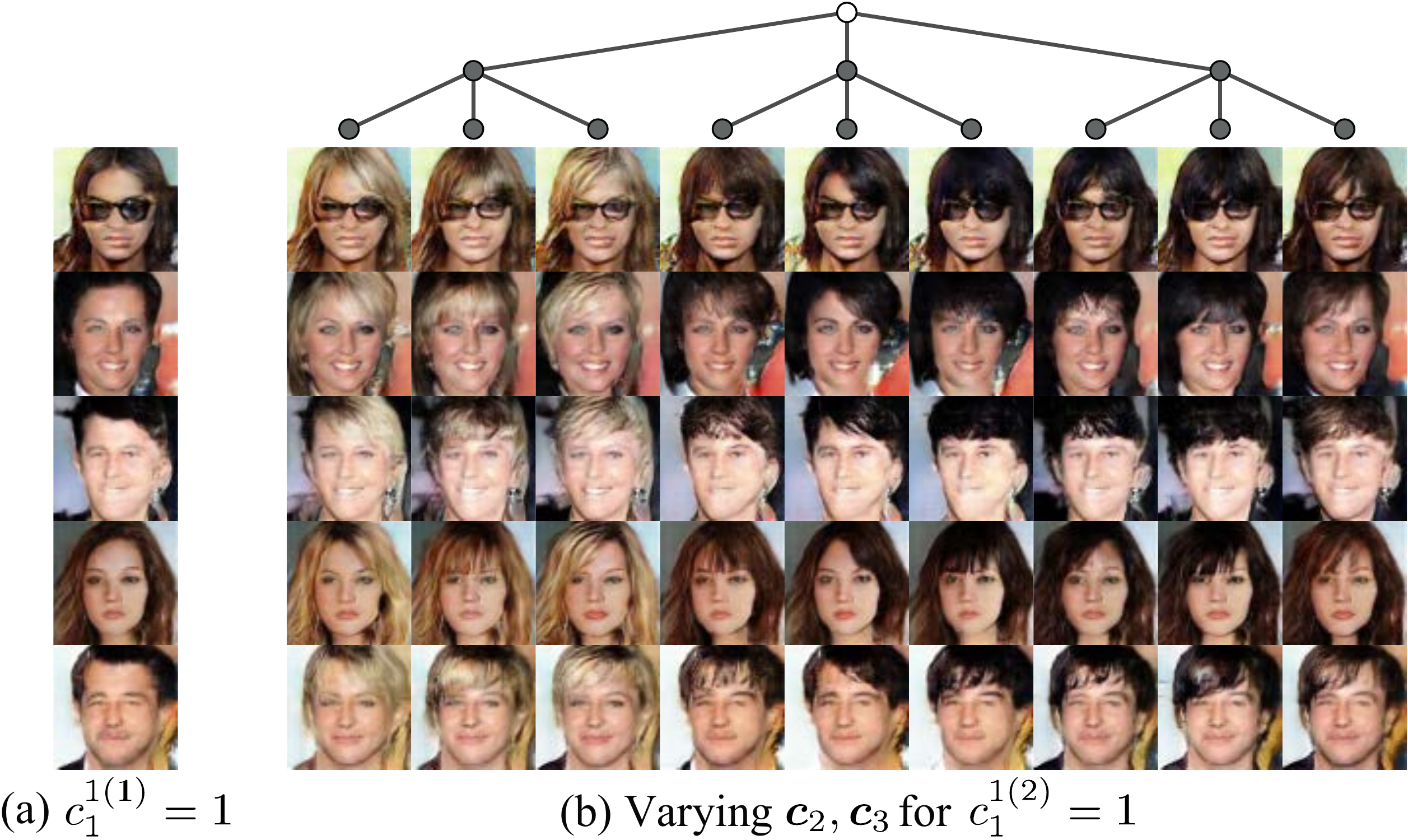}
\end{center}
\caption{{\bf Manipulating latent codes in
    DTLC$^3$-GAN$_{\rm WS}$ on CelebA (bangs):}
  Each column includes five samples generated from
  same ${\bm c}_1$, ${\bm c}_2$, and ${\bm c}_3$ but different ${\bm z}$.
  In (a), samples are generated from $c_1^{1(1)} = 1$,
  i.e., attribute is absent.
  In this case, hierarchical representations are not learned.
  In (b), samples are generated from $c_1^{1(2)} = 1$,
  i.e., attribute is present.
  In this case, hierarchical representations are learned.
  In each row, ${\bm c}_2$ and ${\bm c}_3$ are varied
  per three images and per image, respectively.}
\label{fig:celeba_bangs}
\end{figure*}

\begin{figure*}
\begin{center}
  \includegraphics[width=0.55\textwidth]{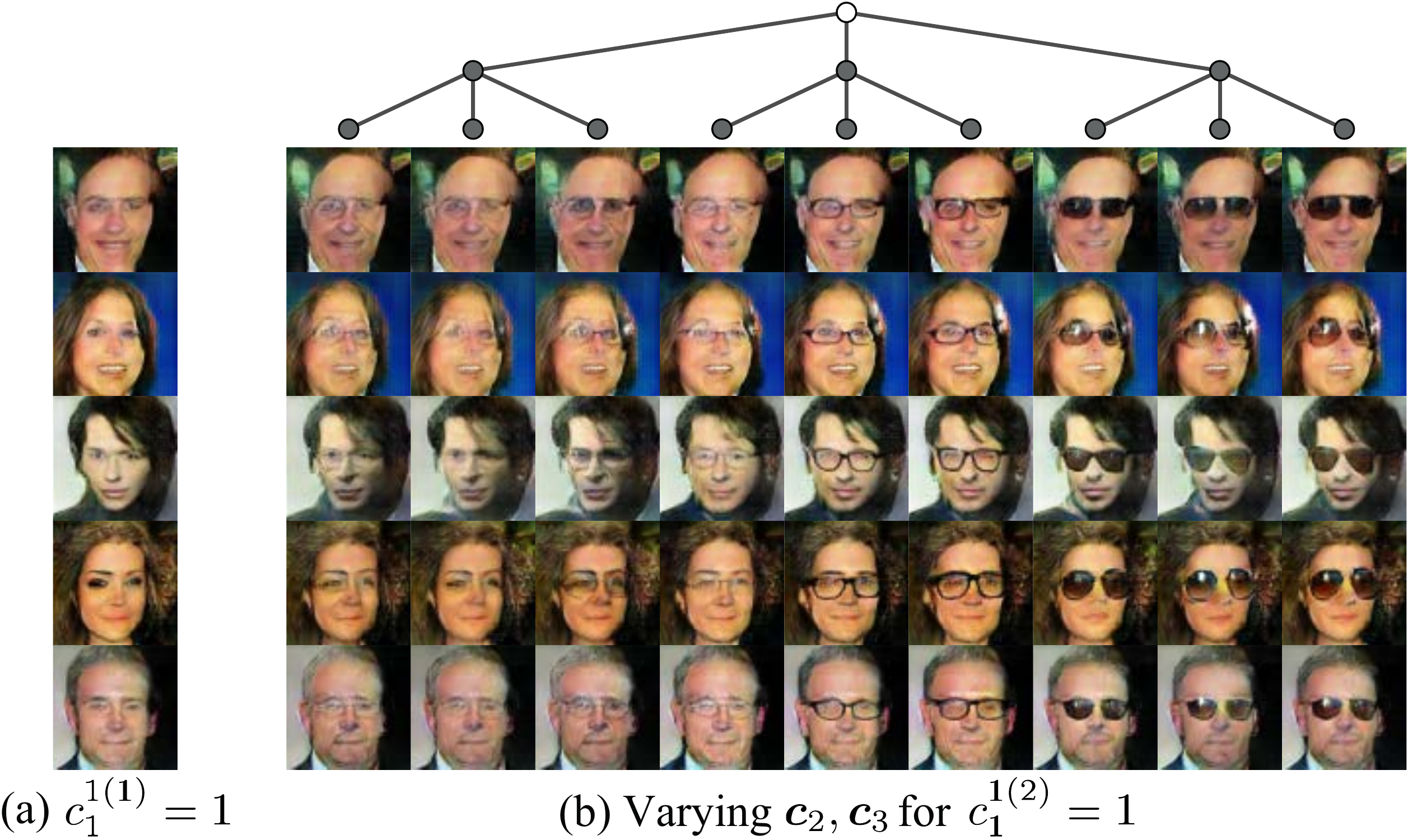}
\end{center}
\caption{{\bf Manipulating latent codes in
    DTLC$^3$-GAN$_{\rm WS}$ on CelebA (glasses):}
  View of figure is same as that in Figure~\ref{fig:celeba_bangs}}
\label{fig:celeba_glasses}
\end{figure*}

\begin{figure*}
\begin{center}
  \includegraphics[width=0.55\textwidth]{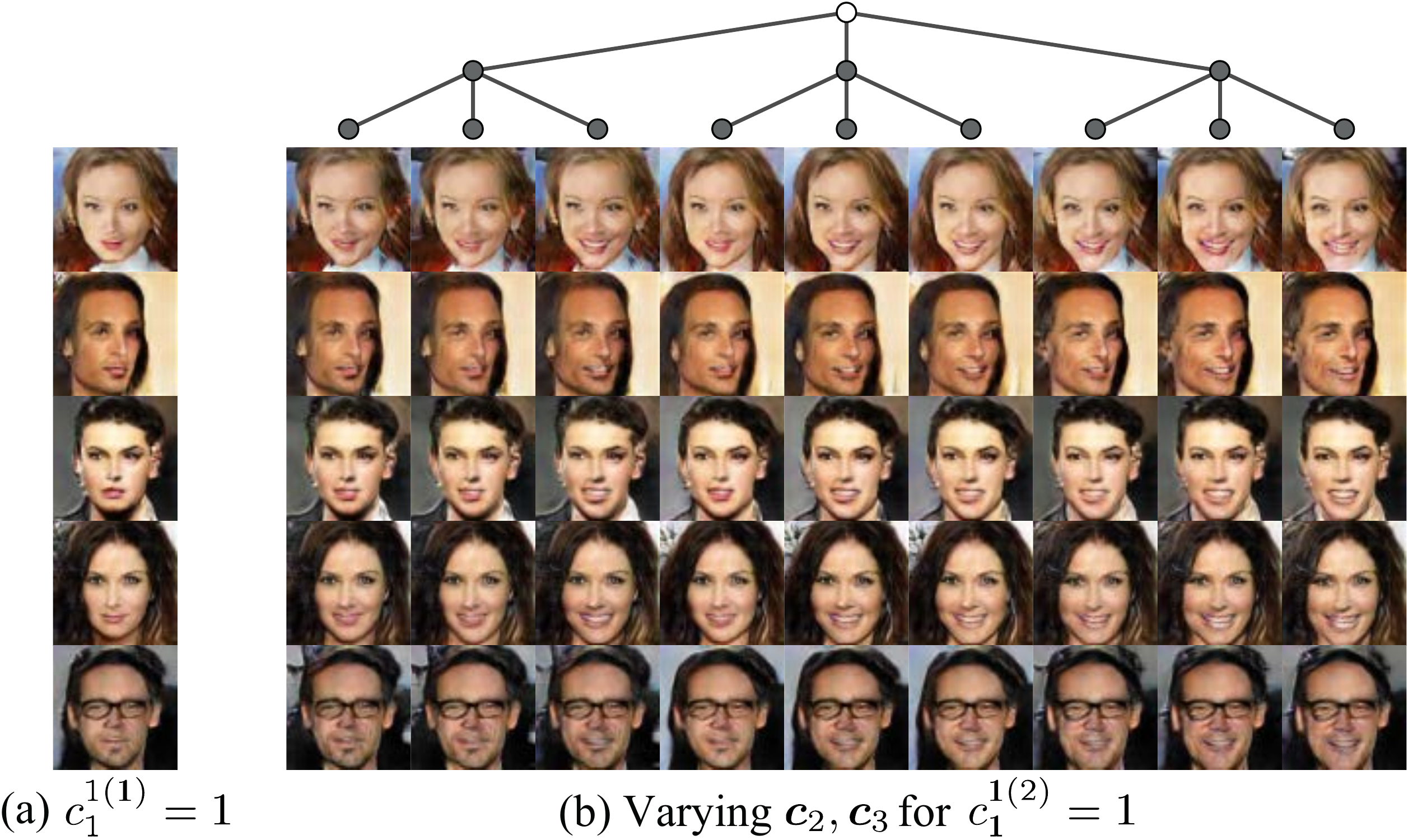}
\end{center}
\caption{{\bf Manipulating latent codes in
    DTLC$^3$-GAN$_{\rm WS}$ on CelebA (smiling):}
  View of figure is same as that in Figure~\ref{fig:celeba_bangs}}
\label{fig:celeba_smiling}
\end{figure*}

We show the generated image examples
using the models
discussed in Section~\ref{subsec:retrieval},
in Figures~\ref{fig:celeba_bangs}--\ref{fig:celeba_smiling}.
We used the {\bf DTLC$^3$-GAN$_{\rm WS}$},
where $k_1 = 2$ and $k_2, k_3 = 3$,
and particularly hierarchical representations are learned only for
the attribute presence state.
We show the results for bangs, glasses, and smiling
in Figures~\ref{fig:celeba_bangs}, \ref{fig:celeba_glasses},
and \ref{fig:celeba_smiling}, respectively.
These results indicate that
the DTLC-GAN$_{\rm WS}$ can learn
attribute-specific hierarchical interpretable representations
by only using the supervision of the binary indicator of attribute presence.

\begin{figure*}
\begin{center}
  \includegraphics[width=\textwidth]{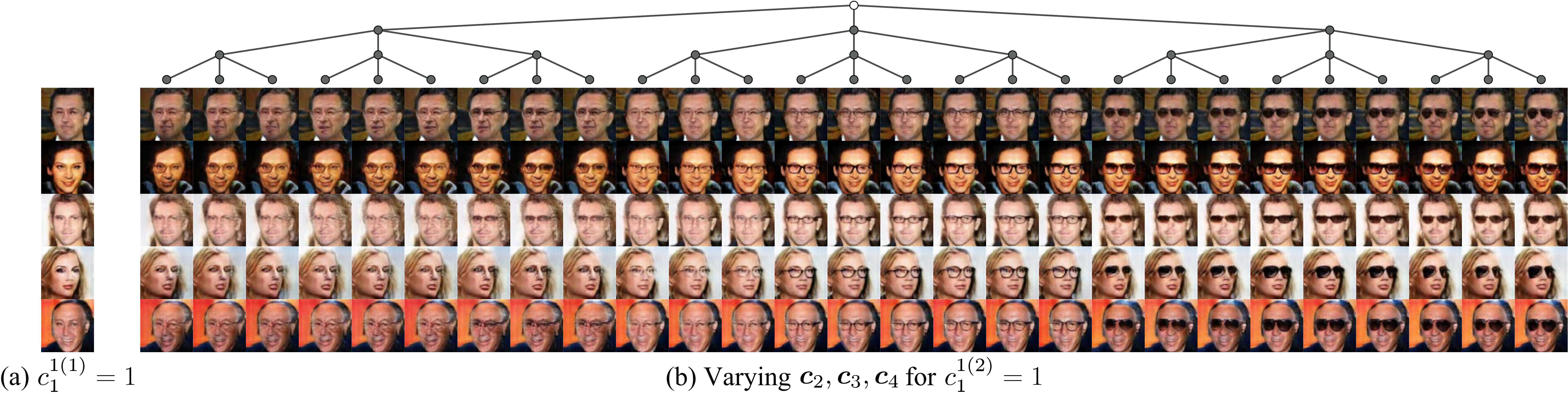}
\end{center}
\caption{{\bf Manipulating latent codes in
    DTLC$^4$-GAN$_{\rm WS}$ on CelebA (glasses):}
  View of figure is similar to that in Figure~\ref{fig:celeba_bangs}.
  Total of $1 + 1 \times 3 \times 3 \times 3 = 28$ categories were learned
  in weakly supervised settings.}
\label{fig:celeba_glasses_h4}
\end{figure*}

\begin{figure*}
\begin{center}
  \includegraphics[width=0.8\textwidth]{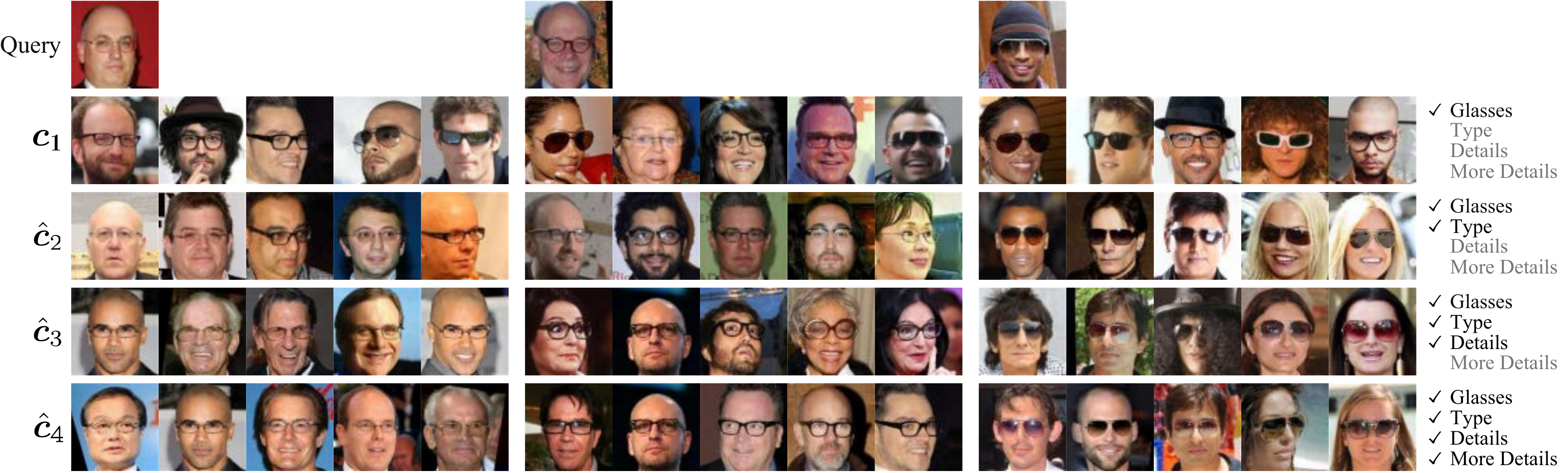}
\end{center}
\caption{{\bf Hierarchical image retrieval using DTLC$^4$-GAN$_{\rm WS}$
    on CelebA (glasses)}}
\label{fig:celeba_glasses_ret_h4}
\end{figure*}

We also show the generated image examples
and image retrieval examples using the {\bf DTLC$^4$-GAN$_{\rm WS}$},
where $k_1 = 2$ and $k_2, k_3, k_4 = 3$,
in Figures~\ref{fig:celeba_glasses_h4}~and~\ref{fig:celeba_glasses_ret_h4},
respectively.
We show the results for glasses.
In this model,
a total of $1 + 1 \times 3 \times 3 \times 3 = 28$ categories were learned
in a weakly supervised setting.
These results indicate that
more detailed semantic features were captured in the lower layers.
As quantitative evaluation of image retrieval,
we calculated attribute-specific SSIM scores.
The scores for 
${\bm c}_1$, $\hat{\bm c}_2$, $\hat{\bm c}_3$, and $\hat{\bm c}_4$
were 0.188, 0.257, 0.266, and 0.267, respectively.
These results indicate that as the layer becomes lower,
the correspondence rate in attribute-specific areas becomes larger and
support the qualitative observations.

\end{document}